\documentclass[runningheads]{llncs}

\usepackage{eccv}

\usepackage{eccvabbrv}

\usepackage{graphicx}
\usepackage{booktabs}

\usepackage{lmodern}  

\usepackage{soul}
\usepackage[T1]{fontenc}  
\usepackage[cmintegrals]{newtxmath}
\usepackage{bm}
\usepackage{amsmath}

\usepackage{xspace} 
\usepackage{multirow} 

\usepackage{graphicx}
\graphicspath{{figures/}}

\usepackage{relsize}
\usepackage{nicematrix}

\usepackage{array}

\usepackage{makecell}

\usepackage{url}

\newcommand{\byteformer}{BUSCA\xspace}

\newcommand{\byteformeracro}{\textbf{B}uilding \textbf{U}nmatched trajectorie\textbf{S} \textbf{C}apitalizing on \textbf{A}ttention}

\newcommand{\sts}{STE}

\definecolor{exemplargreen}{HTML}{f3d6d7}
\definecolor{searchred}{HTML}{f3d6d7}
\definecolor{darkgreen}{HTML}{445b00}
\definecolor{lightergray}{HTML}{dddddd}
\definecolor{goodgreen}{rgb}{0.0, 0.56, 0.0}
\definecolor{badgray}{HTML}{666666}

\definecolor{buscasota}{rgb}{0.98, 0.94, 1.0}

\newcommand{\gooddelta}[1]{\scalebox{0.9}{($\bm{\textcolor{goodgreen}{#1}}$)}}
\newcommand{\baddelta}[1]{\scalebox{0.9}{($\bm{\textcolor{badgray}{#1}}$)}}

\definecolor{rankacolor}{HTML}{bebeff}
\definecolor{rankbcolor}{HTML}{d9d9ff}
\definecolor{rankccolor}{HTML}{e9e9ff}

\newcommand{\rankA}[1]{\textbf{#1}}
\newcommand{\rankB}[1]{#1}
\newcommand{\rankC}[1]{#1}

\newcommand{\plusours}{ \ $\mathbf{+}$ \textbf{\byteformer}}

\newcommand{\icon}[1]{\includegraphics[height=10pt]{#1}}

\usepackage[accsupp]{axessibility}  

\usepackage{hyperref}

\usepackage{orcidlink}

\begin{document}

\title{Lost and Found: Overcoming Detector Failures in Online Multi-Object Tracking}

\titlerunning{Lost and Found: Overcoming Detector Failures in Online MOT}

\author{Lorenzo~Vaquero\inst{1,2}\orcidlink{0000-0002-1874-3078} \and
Yihong~Xu\inst{3}\orcidlink{0000-0003-1043-0656} \and
Xavier~Alameda-Pineda\inst{4}\orcidlink{0000-0002-5354-1084} \and
V\'{i}ctor~M.~Brea\inst{2}\orcidlink{0000-0003-0078-0425} \and
Manuel~Mucientes\inst{2}\orcidlink{0000-0003-1735-3585}}

\authorrunning{L.~Vaquero et al.}

\institute{Fondazione Bruno Kessler, Italy\\
\email{lvaquerootal@fbk.eu} \and
CiTIUS, Univ. of Santiago de Compostela, Spain\\
\email{\{victor.brea, manuel.mucientes\}@usc.es} \and
Valeo.ai, France\\
\email{yihong.xu@valeo.com} \and
Inria Grenoble, Univ. Grenoble Alpes, France\\
\email{xavier.alameda-pineda@inria.fr}}

\maketitle

\begin{abstract}

Multi-object tracking (MOT) endeavors to precisely estimate the positions and identities of multiple objects over time.
The prevailing approach, tracking-by-detection (TbD), first detects objects and then links detections, resulting in a simple yet effective method.
However, contemporary detectors may occasionally miss some objects in certain frames, causing trackers to cease tracking prematurely.
To tackle this issue, we propose \byteformer, meaning `to search', a versatile framework compatible with any online TbD system, enhancing its ability to persistently track those objects missed by the detector, primarily due to occlusions.
Remarkably, this is accomplished without modifying past tracking results or accessing future frames, i.e., in a fully online manner.
\byteformer generates proposals based on neighboring tracks, motion, and learned tokens.
Utilizing a decision Transformer that integrates multimodal visual and spatiotemporal information, it addresses the object-proposal association as a multi-choice question-answering task.
\byteformer is trained independently of the underlying tracker, solely on synthetic data, without requiring fine-tuning.
Through \byteformer, we showcase consistent performance enhancements across five different trackers and establish a new state-of-the-art baseline across three different benchmarks.
Code available at: \url{https://github.com/lorenzovaquero/BUSCA}.
\keywords{2D Tracking \and Multi-target tracking \and Online}
\end{abstract}
  
\begin{figure}[!t]
\centering
\includegraphics[width=0.75\linewidth]{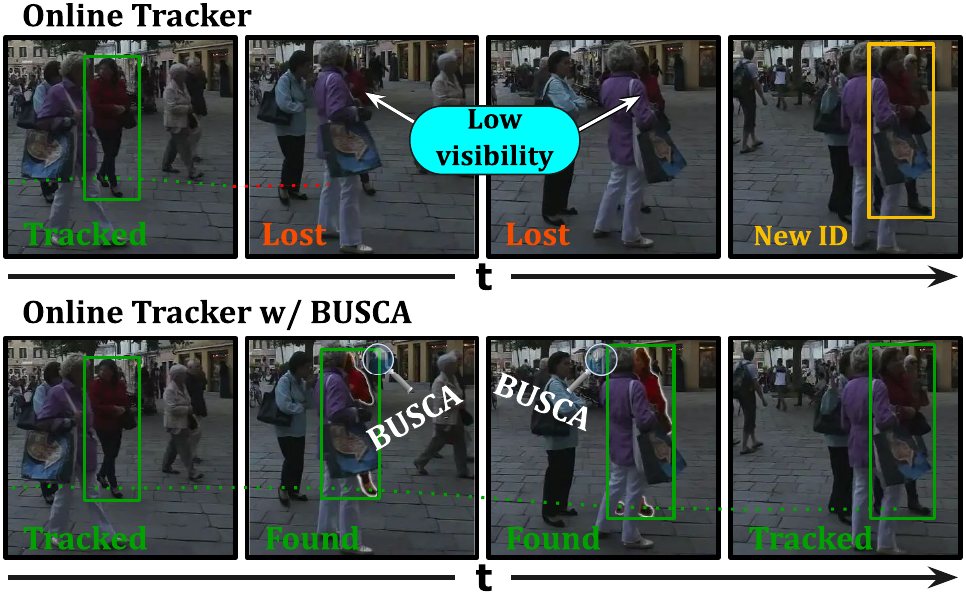}
\caption{{Due to occlusions, detectors fail to locate many relevant elements on a scene (e.g., the woman in red). Accordingly, online multi-object trackers may {lose track of} some objects. 
With \byteformer, we propose a fully online framework that can be integrated into any online TbD tracker to persistently track those objects missed by the detector.} Box colors represent object identities.}
\label{fig:teaser-qualitative}
\end{figure}

\section{Introduction\label{sec:byteformer-introduction}}
Multi-object tracking (MOT) entails the process of locating and identifying multiple objects over time within a scene. It is a {crucial} task in computer vision with applications spanning various domains such as robotics~\cite{Gad2022}, autonomous vehicles~\cite{Guo2022, xu2024towards}, and video surveillance systems~\cite{Rani2023}.
The prevalent MOT paradigm is tracking-by-detection (TbD)~\cite{Dai2022}, {where object trajectories are obtained by (i) first detecting objects and (ii) then associating detections. Although alternative frameworks have been proposed in the literature \cite{Meinhardt2021, Bergmann2019},
TbD has surfaced capitalizing on significant progress in object detection.}
Notably, over the past few years, center- \cite{Zhou2020b, Xu2023} and Transformer-based architectures \cite{Xu2023, Sun2020b} have emerged.
More recently, the MOT performance has been further improved thanks to the adoption of YOLO-based detectors~\cite{Redmon2016, Ge2021} coupled with a straightforward intersection-over-union (IoU) matching.
This simple yet effective approach has even contributed to the renewed popularity of SORT~\cite{Redmon2016, Cao2022_OCSORT,du2023strongsort}.

Meanwhile, significant efforts in the community have been also dedicated to improving identity consistency within a trajectory. This is achieved by devising better association schemes \cite{Zhang2022,Cao2022_OCSORT,du2023strongsort, Zhou2022b} or through re-identification (Re-ID) \cite{Seidenschwarz2022,ren2023focus}. However, these methods remain highly dependent on the availability of detections, which makes them susceptible to trajectory fragmentations.

Current state-of-the-art detectors are not perfect and fail to detect all the objects in a video.
{To have an idea, 17\% of the detections in MOT17~\cite{Milan2016a} validation set are still missed by the YOLOX detector~\cite{Ge2021}, and the extremely occluded objects ($\text{visibility} = 0$, provided in the ground-truth annotations) contribute 11.0 points to the MOTA score based on the standard MOT evaluation \cite{Milan2016a, Dendorfer2020}.}
Meanwhile, modern online trackers pause or terminate the tracking process during these situations where an object fails to be detected, leading to suboptimal results.
We argue that more care should be taken in this regard, avoiding premature terminations of objects that genuinely exist. In this work, we introduce \textbf{\byteformer} (\byteformeracro), which helps online TbD systems handle those objects, often highly occluded, overlooked by the detector.
\byteformer propagates unmatched tracks and, by design, can be applied to the outcome of any online TbD track assignment process.

Some works in the literature~\cite{Saleh2021, dendorfer2022quo, qin2023motiontrack} focus on repairing fragmented tracks and improving trajectory continuity.
However, these have so far been implemented through offline methods, as they alter decisions made on previous time steps (e.g., interpolating a trajectory after re-detection) and/or leverage future information.
Thus, despite some of them claiming to be online, they should be considered as offline according to the widely accepted definition of 'online' in MOTChallenge~\cite{Milan2016a, Dendorfer2020} where \emph{``the solution has to be immediately available with each incoming frame and cannot be changed at any later time''}. The offline fashion makes them impractical for certain real-world applications and not comparable to online methods.
Conversely, \byteformer is able to \emph{persistently track undetected objects in a fully online setting}\footnote{\byteformer strictly respects the `online' definition, thus `fully online'.}.

As an example illustrated in \cref{fig:teaser-qualitative}, some objects are missed due to low visibility even by a highly performant detector~\cite{Ge2021}, causing the tracker to lose them.
With \byteformer, we can enhance any TbD online tracker to continuously track those undetected objects without resorting to offline methods.
To this end, \byteformer is built on a multi-choice question-answering Transformer that finds undetected objects given {(i) \textit{candidate} generated with a motion model (independent of the detector), (ii) \textit{contextual information} derived from neighboring objects, and (iii) \textit{previous observations} from the object of interest.
These inputs are composed of visual and spatiotemporal information.
The visual component characterizes object appearances while the spatiotemporal element encapsulates the size, center location, and timing of the object in a condensed format using an innovative spatiotemporal encoder.

{In summary, the main contributions and novelties of this work are as follows:
\begin{itemize}

  \item \byteformer is a \emph{general} framework to persistently track those objects missed by the detector, in a fully online manner, without (i) modifying past tracking predictions (ii) or accessing future frames.

  \item \byteformer entails (i) a novel \textit{Decision Transformer} inspired by multi-choice question-answering tasks, (ii) a \textit{Proposal Generator} that relies on neighboring tracks, motion, and learned tokens, and (iii) an innovative \textit{Spatiotemporal Encoder} that captures the size, location, and time of the objects.
  The network is trained independently from the underlying tracker and using synthetic data~\cite{Fabbri2021}, without any fine-tuning on real MOT sequences.
  
  \item \byteformer can be seamlessly integrated on top of any online TbD tracker, as demonstrated in our comprehensive experiments where we systematically enhance the performance of five distinct trackers on standard benchmarks~\cite{Milan2016a, Dendorfer2020}, defining a new state-of-the-art among online trackers.
\end{itemize}
}

\section{Related Work\label{sec:related_work}}

\textbf{{End-to-end MOT}} methods model detection, tracking, and their implicit matching within a unified architecture.
The most common approaches tackle this through identity embeddings~\cite{Wang2020}, regression~\cite{Bergmann2019,TbD} or the recent use of attention mechanisms~\cite{Meinhardt2021, Zhu2021, Zeng2022, Zhou2022b, Gao_2023_ICCV, Cai2022}.
Nonetheless, this holistic design can create challenges during the joint training process~\cite{Gao_2023_ICCV} and, prevent these methods from being applicable to other trackers and leveraging leading-edge detectors.
Consequently, these models have {not} yet superseded TbD techniques.

\noindent \textbf{Tracking by detection (TbD)} is an effective paradigm that decouples the MOT task into object detection and data association. This decomposition enables TbD methods~\cite{Xiang2015,Zhao2020,Zhou2020b,He2021, Zhang2022,Seidenschwarz2022,Sun2020b,Xu2023} to benefit from classical~\cite{Xiang2015,Redmon2016,Ren2017}, more advanced~\cite{Zhang2022, Khan2022} or self-constructed~\cite{Sun2020b,Xu2023,Zhou2020b} detectors, coupled with diverse association processes such as hierarchical clustering~\cite{Zhao2020}, graph neural networks~\cite{He2021} or geometric cues~\cite{Zhang2022}.

{In particular, center-based methods like CenterTrack \cite{Zhou2022b} and TransCenter \cite{Xu2023} alleviate the ambiguity in bounding boxes by predicting object center heatmaps in a CNN-based or Transformer-based architecture, respectively.} Recently, ByteTrack \cite{Zhang2022} showcases remarkable results using a meticulously tuned YOLOX detector~\cite{Ge2021} paired with a simple IoU-based matching mechanism. This powerful detector has also revived SORT~\cite{Bewley2016} with a stronger association mechanism in methods such as OC-SORT and StrongSORT~\cite{Cao2022_OCSORT, du2023strongsort}. 
Nevertheless, these TbD trackers remain highly vulnerable to missed detections.
{This issue motivates us to introduce \byteformer, a framework designed to improve any online TbD tracker by persistently tracking those objects overlooked by the detector.}

\noindent {\textbf{Improving trajectory consistency}, i.e., maintaining consistent object identities over time, is one of the main challenges of online multi-object trackers.
Most of these methods rely on frame-by-frame association of {detections} solved via Hungarian matching~\cite{Kuhn1955}. However, pure motion-based associations~\cite{Bewley2016, Bochinski2017, Zhang2022} often encounter difficulties in crowded environments or moving-camera scenarios.
As a result, other works turn to appearance-based techniques~\cite{Wojke2017,Kim2021,Shuai2021,Vaquero2021,ren2023focus,rafi2020self}, hybrid cues~\cite{Seidenschwarz2022,Vaquero2023,du2023strongsort,TBMotion}, or Transformer solvers~\cite{Zhou2022b,Zeng2022}. {Notably, GHOST \cite{Seidenschwarz2022} redesigns the use of a ReID model and builds a simple yet strong baseline.} In efforts to lessen the impact of occlusions, some methods aim to predict an object's visibility in order to adjust its detections' confidences~\cite{Jiang2022} or re-weight the association matrix~\cite{Zhang2022b}. On the other hand, some strategies improve associations by hallucinating object trajectories~\cite{Tokmakov2021} or by prompting re-detections in areas where occluders are present~\cite{Liu2022}.

Nonetheless, unlike \byteformer, these more advanced association processes \emph{remain heavily dependent on the detector as they operate on available detections.} \cite{TBMotion} is a rare exception but at the cost of MOT performance drop.

\noindent {\textbf{Ensuring trajectory continuity} is a non-trivial task that attempts to repair the trajectory of an object from the instant it is lost until it is re-identified again.
Thus, most current trackers perform an extra \emph{offline} post-processing step based on linear~\cite{Zhang2022} or Gaussian-smoothed~\cite{du2023strongsort} interpolation.
Some more sophisticated methods involve implementing a probabilistic model to retroactively insert missed detections~\cite{Saleh2021}, learning an additional Refind Module~\cite{qin2023motiontrack} to bridge these gaps, or 2D-to-3D lifting and performing motion forecasting in a bird's eye view~\cite{dendorfer2022quo}.
Nevertheless, these strategies remain \textit{offline}~\cite{Milan2016a, Dendorfer2020} as they either alter predictions on past time steps or take into account future frames, limiting their applicability in certain real-world scenarios. We introduce thus \byteformer, a framework that can be \emph{built on top of any online TbD tracker} to enhance its continuity and consistency \emph{in a fully online fashion}.

\begin{figure*}[ht]
\centering
\includegraphics[width=\linewidth]{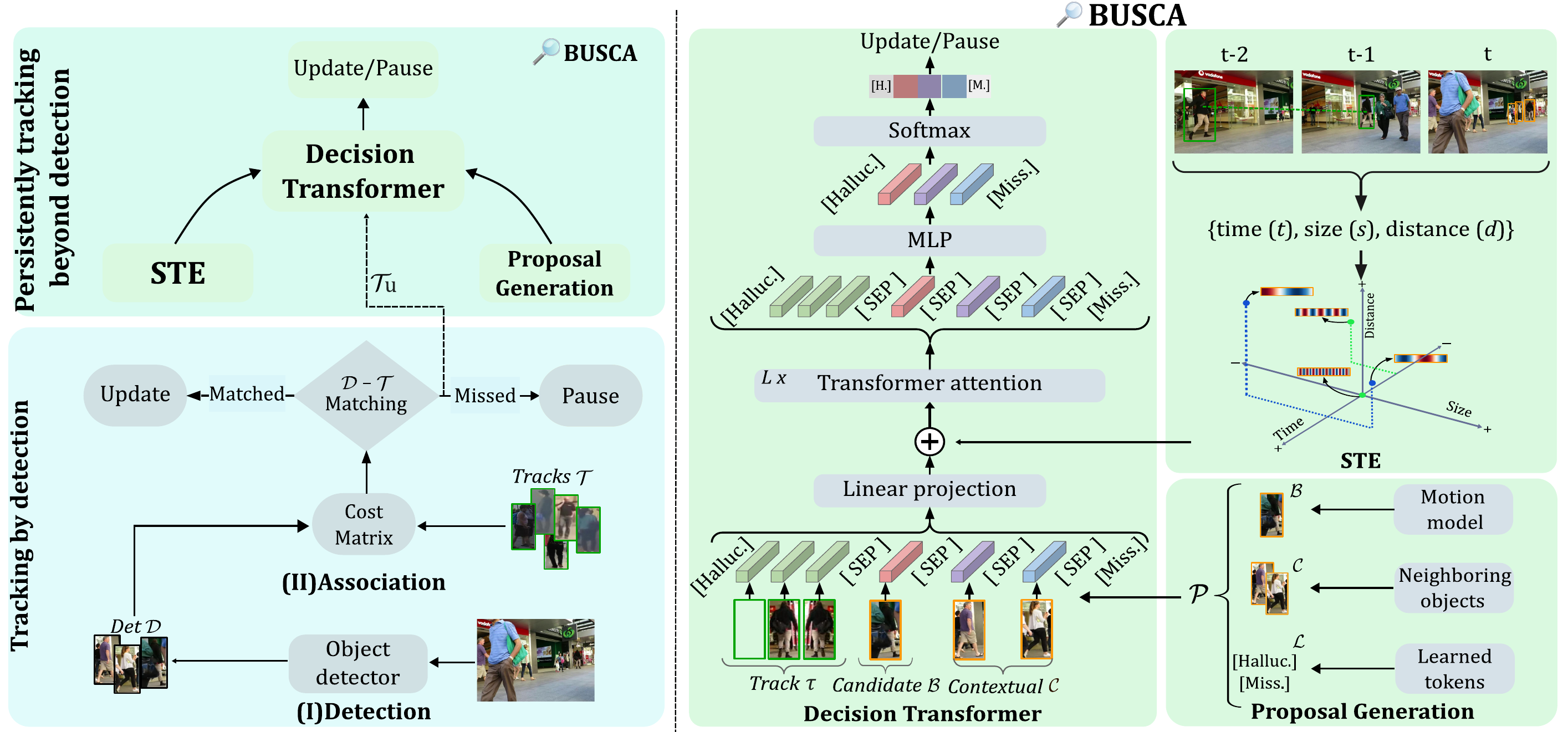}
\caption{
{The bottom-left panel depicts the tracking-by-detection (TbD) paradigm (\cref{sec:tbd_explanation}), where a track is paused when the detector fails to locate the object.
To address this issue, we integrate \byteformer into the online TbD tracker (\cref{sec:busca}) as shown in the top-left panel.
This allows for the extension of trajectories of undetected objects by pairing them with proposals comprising \textit{candidates} ($\mathcal{B}$), \textit{contextual information} ($\mathcal{C}$) and \textit{learned tokens} ($\mathcal{L}$) (\cref{sec:method-candidates}) via an innovative decision Transformer (\cref{sec:method-byteformer}).
Comprehensive details about the components of \byteformer are showcased in the right-hand panel.
The \textit{track} observations and proposals fed to the decision Transformer are made up of both appearance features (extracted with a convolutional backbone omitted here for clarity) and spatiotemporal cues for time, size, and distance encoded in a compact embedding through our novel spatiotemporal encoding (\sts, \cref{sec:method-encoding}).}
}
\label{fig:byte_third}
\end{figure*}

\section{{TbD in a Nutshell}}\label{sec:tbd_explanation}
In the tracking by detection (TbD) paradigm, at a given frame a detector first produces a set $\mathcal{D} = \{\delta_1, ..., \delta_M\}$ of $M$ detections, with each detection $\delta_i = \{a_{i}, c_{i}, \omega_{i}\}$ is defined by its appearance $a_{i}$ (i.e., features of the image contained in the coordinates), coordinates $c_{i}$ (object size and center location) and confidence score $\omega_{i}$. These detections are used to propagate the position of a set $\mathcal{T} = \{\tau_1, ..., \tau_N\}$ of $N$ active tracks, each represented by a time-ordered set $\tau_j = (o_{j,1}, \ldots, o_{j,Z})$ of observations $o_k = \{a_{k}, c_{k}\}$ over the past $Z$ frames.

$\mathcal{D}$ is compared with $\mathcal{T}$, using coordinates and geometric cues~\cite{Bewley2016,Zhang2022}, appearance information~\cite{Wojke2017}, or both~\cite{Seidenschwarz2022}, yielding a cost matrix of size $N \times M$ whose optimal assignments are determined through Hungarian matching~\cite{Kuhn1955}.
Thus, as shown in the bottom-left part of \cref{fig:byte_third}, correctly matched tracks are updated with the assigned detections, while those without a matching detection are paused.
Having correct and sufficient detections for all tracks is critical, leading many trackers to resort to offline interpolation techniques to repair missing observations.
In order to address this issue without resorting to offline interpolation, we present \byteformer, which tracks those undetected objects in a fully online fashion.

\section{\texorpdfstring{\protect\icon{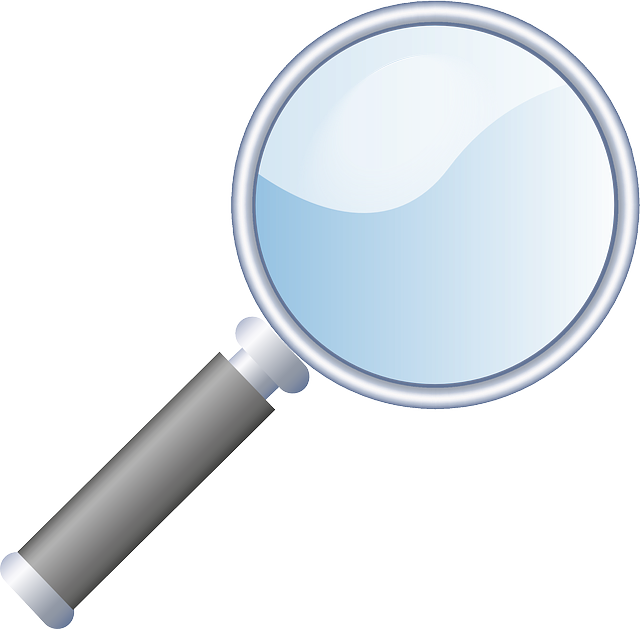}}{} \byteformer: Finding Objects without Detections} \label{sec:busca}
{Current detectors still fail to detect all the objects, especially in low-visibility situations i.e., heavy occlusions.
Modern trackers heavily rely on the detection quality, thus naively stopping the tracking process whenever the detector fails. Therefore, \byteformer comes to help by saving those objects missed by the detector and finding where they are.}

In particular, \byteformer is a fully online framework that can be coupled with any TbD tracker to persistently track those objects missed by the detector.
As can be seen in the upper left part of \cref{fig:byte_third}, \byteformer receives unmatched tracks $\mathcal{T}_u$ and compares them with a set of proposals generated through a proposal generation process (\cref{sec:method-candidates}).
This comparison is carried out through a novel decision Transformer (\cref{sec:method-byteformer}), which uses an innovative spatiotemporal encoding (\sts, \cref{sec:method-encoding}) to aggregate information of different nature.
This way, \byteformer can update the coordinates of those unmatched tracks or determine whether they have really left the scene.

\subsection{Decision Transformer: To Be or Not To Be} \label{sec:method-byteformer}
Deciding whether to pause an undetected track or propagate its identity can be formulated as a multiple-choice question-answering task~\cite{Radford2018}.
That is, given a question (the track $\tau$) and a set of possible options (the proposals $\mathcal{P} = \{p_{1}, ..., p_{J}\}$, where $p_{i} = \{a_{i}, c_{i}\}$), the goal of the network is to find the correct answer (the decision of which proposal to match to the track) forming the assignment set $\mathcal{A} = \{\tau_j \mapsto p_{i} | \tau_j \in \mathcal{T}, p_i \in \mathcal{P}\}$.
Inspired by this formulation, we propose to maintain undetected objects via a Transformer-based design that inputs different \textit{proposals} and a \textit{track}, outputting the best match, i.e., the proposal with the highest probability.

As shown on the right side of \cref{fig:byte_third}, our decision Transformer is implemented through an $L$-layer encoder model, which receives an input $\mathcal{I} = \{\tau, \mathcal{P}\}$, in which the past observations of the track are included.
For each of the individual elements that make up the input (referred to as \emph{tokens}), the appearance information $a$ is processed by a convolutional backbone and projected to a lower dimensional space.
This visual information of each token is then fused with its geometric cues $c$ using our innovative spatiotemporal encoding (\cref{sec:method-encoding}), to allow the Transformer to reason complex relationships between motion and visual features.

Within the decision Transformer, the input tokens are self-attended with each other, yielding refined tokens $\mathcal{J} = \{\overline{\tau}, \overline{\mathcal{P}}\}$ where the features most closely related to the track have been enhanced.
Then, the elements of $\overline{\mathcal{P}}$ are fed to a shared-weight multi-layer perceptron (MLP) that generates one logit per token.
After a Softmax operation, we output the probabilities that the track $\tau$ is assigned to each proposal $p$, allowing us to obtain $\mathcal{A}$ by finding the maximum probability.
Finally, we update $\tau$ when it is successfully matched with a candidate proposal (See \cref{sec:method-candidates}) or pause it otherwise.
It should be noted that the MLP is share-weight, so as not to be restricted to any fixed input size.

\subsection{Proposal Generation: Missing Puzzle Pieces\label{sec:method-candidates}}
As with textual question-answering problems, the composition of the proposals $\mathcal{P}$ is one of the most critical aspects, and this is no different for our decision Transformer.
$\mathcal{P} = \{\mathcal{B}, \mathcal{C}, \mathcal{L}\}$ is composed of candidates $\mathcal{B}$, contextual proposals $\mathcal{C}$, and learned proposals $\mathcal{L}$.
As shown in the bottom-right of \cref{fig:byte_third}, $\mathcal{B}$ and $\mathcal{C}$ are extracted from the frame, while $\mathcal{L}$ is learned. \byteformer will keep a track $\tau$ active and update it with the proposal information if it is associated with any element from $\mathcal{B}$ and pause $\tau$ otherwise.

Generating the sets of proposals $\mathcal{B}$ and $\mathcal{C}$ is nontrivial given that none of the detections in $\mathcal{D}$ can be associated with $\tau$.
Given its reasonable performance~\cite{Bewley2016, Zhang2022,du2023strongsort}, we opt for a simple yet effective Kalman filter~\cite{Kalman1961} to predict a new observation of $\tau$ at the current frame.
To this end, it is possible to obtain $\mathcal{B} = \{\text{Kalman}(\tau)\}$ without adding extra complexity to \byteformer, all while effectively managing complex motion scenarios, as evidenced in the supplementary material.
Regarding the contextual proposals $\mathcal{C}$, their goal is to provide \byteformer with more information about the scene.
$\mathcal{C}$ is composed of the $Q$ closest observations within the neighborhood of $\tau$, $V(\tau)$.
Details for the computation of the maximum neighborhood distance for $\tau$ are given in the supplementary material.

The input proposals $\mathcal{P}$ of \byteformer also comprise a set $\mathcal{L} = \{\text{\texttt{[Halluc.]}}, \text{\texttt{[Miss.]}}\}$ of learned tokens that allow the Transformer to make complex decisions about the tracking process and pause $\tau$ if necessary.
Specifically, \texttt{[Halluc.]} is learned to capture whether any observation $o$ is corrupted (i.e., belonging to a different object) whereas \texttt{[Miss.]} handles if $\tau$ has left the scene or none of the elements of $\{\mathcal{B}, \mathcal{C}\}$ are suitable enough to be matched.
Additionally, a separator token \texttt{[SEP]} borrowed from textual Transformers~\cite{Radford2018} is also learned to delimit each of the elements of $\mathcal{P}$.

\subsection{Spatiotemporal Encoding (\sts): Merging Modalities\label{sec:method-encoding}}
Along with appearance features, spatiotemporal information is also crucial for making correct assignments.
This information is however more complex to be encoded due to its multi-dimensionality (i.e., time-stamp $t$ at which observations are recorded, the size $s$ of the bounding box, and their distance $d$ in the 2D coordinate space).
To this end, we propose the spatiotemporal encoding (\sts) depicted on the top-right part of \cref{fig:byte_third}, which models these relationships between observations and allows its fusion with visual features so \byteformer can effectively learn complex relationships.
Our spatiotemporal encoding supersedes the conventional positional encoding often implemented in Transformer models~\cite{Vaswani2017}.
This encoding is generated through a two-step process comprising the \emph{interplay mapping} and subsequent the \emph{embedding projection}.

{\textbf{Interplay mapping}.
The encodings employed in visual Transformers rely on absolute values, which limit the network's overall adaptability and make them rely on interpolation techniques to handle diverse frame sizes~\cite{Dosovitskiy2021, Caron2021, Oquab2023}.
Moreover, this method has consequential downsides for tracking tasks, as identical interactions might be represented differently depending on their specific occurrence (e.g. proximity between a track and an observation will be encoded differently depending on their absolute position within the frame or video).

To address this, our \sts\ relies on a novel interplay mapping that models interactions relative to an anchor $\kappa$.
In our specific use case, $\kappa = \{x_{\kappa}, y_{\kappa}, w_{\kappa}, h_{\kappa}, t_{\kappa}\}$ corresponds to the coordinates (i.e., object center, width, and height) and time-stamp of the last known observation of the track $o \in \tau$.
To this end, we can compute a spatiotemporal embedding $\{E^{t}, E^{s}, E^{d}\}$ comprising time, size, and distance, respectively, for each token $\iota \in \mathcal{I}$ as:
\begin{align}
&E^{t} = \sigma^t \left( t_\iota - t_\kappa \right) \\
&E^{s} = \sigma^s \left( \log{\left(\frac{w_{\iota}}{w_{\kappa}}\right)} + \log{\left(\frac{h_{\iota}}{h_{\kappa}}\right)} \right)  \\
&E^{d} = \sigma^d \log{\sqrt{\left(\frac{x_{\iota} - x_{\kappa}}{w_{\kappa}}\right)^2 + \left(\frac{y_{\iota} - y_{\kappa}}{h_{\kappa}}\right)^2}}
\end{align}
where $\sigma^t, \sigma^s, \sigma^d$ are scaling factors. 
This relative representation boosts the generalization capacity of \byteformer and improves convergence during training.

{\textbf{Embedding Projection}.
After computing the interplay mapping between input tokens and $\tau$, it is essential to make this representation compatible with both the transformer and the visual features.
However, adding multiple independent sinusoidal functions could lead to potentially ambiguous information, according to \cite{Wang2021}.
To this end, it is necessary to establish a joint spatiotemporal encoding by expanding the function used in \cite{Vaswani2017} to a 3-dimensional space.
Given the Transformer's internal dimension of $D^{\text{Tr}}$ channels, we equally distribute it among the three components of our spatiotemporal embedding $D = D^{\text{Tr}} / 3$.
Therefore, for a given dimension $E^\Delta$ where $\Delta \in \{t, s, d\} $ we can compute its projected embedding $PE^\Delta$:
\begin{align}
\hspace{-2mm}
{PE}_{2i}^\Delta &= \sin{\left( \frac{E^\Delta}{10000^{2i / D}} \right)} & 
{PE}_{2i + 1}^\Delta &= \cos{\left( \frac{E^\Delta}{10000^{2i / D}} \right)}
\end{align}
where $0 \le i < D/2$.
And subsequently concatenate the components of the different dimensions to create our compact spatiotemporal encoding ${\sts} = \left(PE^{t}, PE^{s}, PE^{d}\right)$ for each one of the tokens $\iota \in \mathcal{I}$.

\section{Experimental Results\label{sec:byteformer-experimentation}}
In \cref{subsec:exp_setting}, we clarify the experimental settings along with the used datasets and metrics.
In \cref{sec:ablation}, we validate the necessity of \byteformer compared to the naive solutions and show that it can systematically extend tracks' lifespan, improving trajectory continuity without losing consistency.
Subsequently, we empirically demonstrate the effectiveness of each component of \byteformer and justify its design choices.
Once validated, we show in \cref{subsec:sota} that \byteformer is a plug-and-play component that consistently improves various trackers, setting new state-of-the-art performance in all tested benchmarks compared to other online methods.
Finally, some successful and failure cases are qualitatively shown in \cref{subsec:quali}.

\subsection{Experimental Settings \label{subsec:exp_setting}}
We conduct our experiments on the widely-used MOT16~\cite{Milan2016a}, MOT17~\cite{Milan2016a} and the crowded MOT20~\cite{Dendorfer2020} datasets.
In contrast to other methods, we train \byteformer using solely synthetic data from MOTSynth~\cite{Fabbri2021}, which consists of 764 full-HD videos recorded at 20 fps.
For each training sample, we construct a track of length $Z=11$ and randomly select $5$ objects near $\tau$ to form a proposal set (current observation of $\tau$ is the positive candidate while objects with an overlap smaller than $0.5$ are negatives.
Additionally, we set a $15\%$ probability of not sampling any positives (\texttt{[Miss.]} will be considered the correct option) and a $1\%$ chance of altering observations within $\tau$ (\texttt{[Halluc.]} will be the correct option).
Our training process focuses only on bounding box annotations and does not require any fine-tuning towards particular datasets or tracking systems.
The computational cost of \byteformer is relatively small, with only 8.7M parameters and a runtime of 45ms per frame on a single NVIDIA RTX GPU (when integrated with \cite{Zhang2022}, the whole system runs at roughly 13fps).

For the ablation, we focus on MOT17 with the widely-adopted split~\cite{Zhou2020b, Zhang2022, Seidenschwarz2022} that evenly divides each video sequence into training and validation sets.
Unless otherwise stated, we employ ByteTrack~\cite{Zhang2022} as our baseline tracker due to its state-of-the-art performance, but we remove its offline interpolation and its per-sequence curated thresholds.
For the comparison with the state-of-the-art, we submit our test set results to the MOTChallenge servers and compare our approach with current \emph{online} methods as defined in the challenge~\cite{Milan2016a, Dendorfer2020}.

For evaluation, we report the standard metrics adopted by the MOTChallenge~\cite{Dendorfer2021}. These include MOTA~\cite{bernardin2008evaluating} reflecting the overall performance of a predicted trajectory; the recently introduced HOTA~\cite{Luiten2021} that balances object coverage and identity preservation; IDF1~\cite{ristani2016performance} focusing on association quality; IDentity SWitches (IDSW) to reflect identity consistency; and False Positives (FP) as well as False Negatives (FN) to assess detection performance.
{Additional experiments and implementation details can be found in the supplementary material.}

\subsection{Model Validation and Ablation\label{sec:ablation}}
\noindent\textbf{{Naive} approaches are not enough.} Persistently tracking objects overlooked by the detector is not a trivial task and cannot be achieved with simpler naive approaches. Specifically, ByteTrack \cite{Zhang2022} demonstrates that with a reliable detector, some low-score detections can be leveraged in a second-round association.
One would then expect that \textbf{L}owering the \textbf{D}etection (\textbf{LD}) threshold $\epsilon=0.01$ would provide further benefits during the tracks-detections matching.
Another direct approach similar to \byteformer consists of using a motion model (e.g., Kalman filter) to estimate the track future coordinates and perform an extra round of associations based on motion and geometry cues like \textbf{IoU}.
Alternatively, we also propose an extra recovery round based on \textbf{Mixed} cues (i.e. both IoU and appearance),  as shown important for more robust associations \cite{Wojke2017}.

\begin{table}[t!]
\centering
\caption{Comparison to different simpler solutions on MOT17~\cite{Milan2016a} val set. The difference with the baseline is depicted next to each metric. ByteTrack \cite{Zhang2022} is used as base tracker removing its offline interpolation and per-sequence thresholds, noted with $^{\star}$.
\vspace*{-1.25mm}
}
\resizebox{0.75\linewidth}{!}{%
\begin{NiceTabular}{@{}l|c@{\hskip 3pt}cc@{\hskip 3pt}cc@{\hskip 3pt}cc@{\hskip 3pt}c@{}}
\CodeBefore
   \rowcolors{7}{buscasota}{}
\Body
\midrule \\[-13pt]
& \multicolumn{2}{c}{\textbf{MOTA} $\uparrow$} & \multicolumn{2}{c}{\textbf{HOTA} $\uparrow$} & \multicolumn{2}{c}{\textbf{FN} $\downarrow$} & \multicolumn{2}{c}{\textbf{FP} $\downarrow$} \\[-2pt]
\midrule
\midrule

ByteTrack$^{\star}$  & 76.5 &  & 67.4 & & 9120 &  & \rankA{3410} &  \\
$+$ LD & 75.3 & \baddelta{-1.2} & 65.6 & \baddelta{-1.8} & 8854 & \gooddelta{-266} & 4196 & \baddelta{+786}\\
$+$ IoU  & 75.4 & \baddelta{-1.1} & 67.0 & \baddelta{-0.4} & \rankA{7588} & \gooddelta{-1532} & 5493 & \baddelta{+2083} \\
$+$ Mixed & {76.6} & \gooddelta{+0.1} & \rankA{67.6} & \gooddelta{+0.2} & 8393 & \gooddelta{-727} & 4063 & \baddelta{+653}  \\

$+$ \textbf{\byteformer (ours)}  &  \rankA{77.1} &  \gooddelta{+0.6} & \rankA{67.6} & \gooddelta{+0.2} & 8326 & \gooddelta{-794} & 3889 & \baddelta{+479} \\[-2pt]
\midrule
\end{NiceTabular}}
\label{tab:ablation-byte}
\end{table}

\begin{figure}[t!]
\centering
\begin{subfigure}{0.49\linewidth}
  \includegraphics[trim={0 0 0 4.5mm},clip, width=\textwidth]{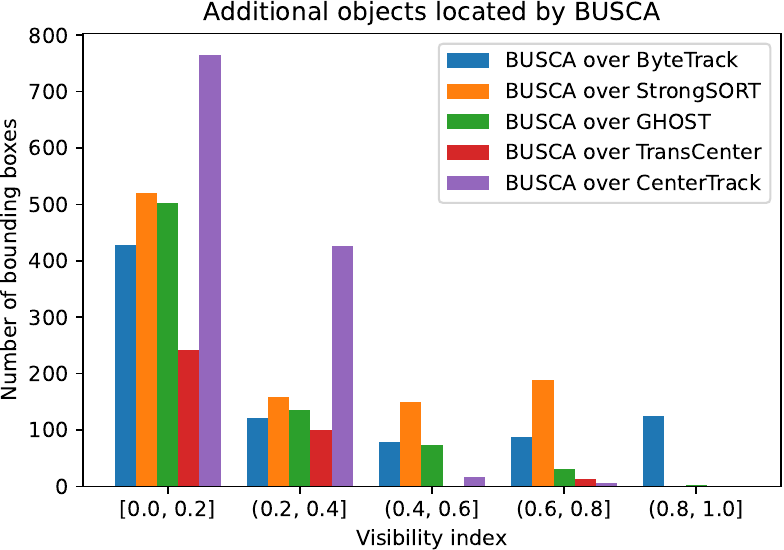}
  \caption{Extra successfully rescued objects.}
\label{fig:ablation-savetracks}
\end{subfigure}
\begin{subfigure}{0.50\linewidth}
  \includegraphics[trim={0 0 0 4.5mm},clip, width=\textwidth]{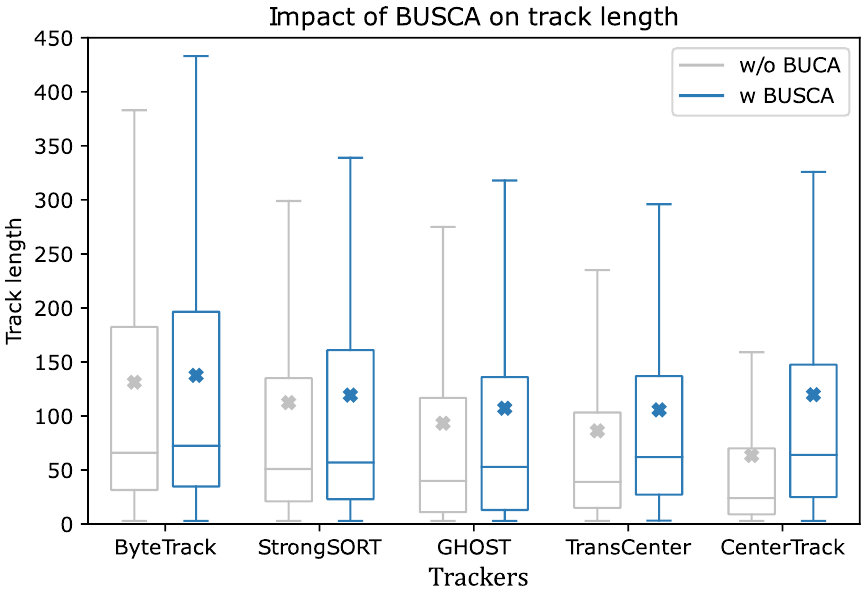}
  \caption{Impact of \byteformer on track length.}
\label{fig:ablation-tracklength}
\end{subfigure}
\caption{(a) Analysis of the additional objects that \byteformer \emph{successfully} locates when integrated with different trackers. The objects are grouped by their visibility~\cite{Milan2016a}. (b) Analysis of the impact of \byteformer on the resulting track length in different trackers. {Additional implementation details can be found in the supplementary material.}
\vspace*{-1mm}}
\end{figure}

\begin{table}[t!]
\centering
\caption{Ablation on MOT17~\cite{Milan2016a} val set of the different components that comprise \byteformer.
HLC=\texttt{[Halluc.]} learned token, MSS=\texttt{[Miss.]} learned token, STE=spatiotemporal encoding, CTX=contextual proposals. 
The difference with the baseline is depicted next to each metric. ByteTrack \cite{Zhang2022} is used as base tracker removing its offline interpolation and per-sequence thresholds.}
\resizebox{0.8\linewidth}{!}{%
\begin{NiceTabular}{@{}c|c|c|c|c|c@{\hskip 3pt}cc@{\hskip 3pt}cc@{\hskip 3pt}cc@{\hskip 3pt}c@{}}
\CodeBefore
   \rowcolors{9}{buscasota}{}
\Body
\midrule \\[-13pt]
Line & HLC & MSS & STE & CTX & \multicolumn{2}{c}{\textbf{MOTA} $\uparrow$} & \multicolumn{2}{c}{\textbf{HOTA} $\uparrow$} & \multicolumn{2}{c}{\textbf{FN} $\downarrow$} & \multicolumn{2}{c}{\textbf{FP} $\downarrow$} \\[-1pt]
\midrule
\midrule
1& & & & & 76.5 & & 67.4 &  & 9120 &  & \rankA{3410} &  \\

2&\checkmark &  &  &  & 75.0 & \baddelta{-1.5} & 66.3 & \baddelta{-1.1} & 8395 & \gooddelta{-725} & 4911 & \baddelta{+1501}  \\
3& & \checkmark &  &  & 76.4 & \baddelta{-0.1} & 67.3 & \baddelta{-0.1} & \rankA{8064} & \gooddelta{-1056} & 4513 & \baddelta{+1103}\\
4&\checkmark & \checkmark &  &  & 76.5 & \baddelta{\ 0.0\ } & 67.1 & \baddelta{-0.3} & 8656 & \gooddelta{-464} & 3853 & \baddelta{+443}  \\
5&\checkmark & \checkmark & \checkmark &  & 76.7 & \gooddelta{+0.2} & 67.4 & \baddelta{\ 0.0\ } & 8528 & \gooddelta{-592} & 3851 & \baddelta{+441} \\
6&\checkmark & \checkmark &  & \checkmark & 76.9 & \gooddelta{+0.4} & 67.6 & \gooddelta{+0.2} & 8387 & \gooddelta{-733} & 3884 & \baddelta{+474}  \\

7& \checkmark & \checkmark & \checkmark & \checkmark & \rankA{77.1} & \gooddelta{+0.6} & \rankA{67.6} & \gooddelta{+0.2} & 8326 & \gooddelta{-794} & 3889 & \baddelta{+479}  \\[-1pt]
\midrule
\end{NiceTabular}}
\label{tab:ablation-byte-components}
\end{table}

As shown in \cref{tab:ablation-byte}, lower-score detections are not reliable and \textbf{+LD} increases FP (+786) with a slight decrease in FN (-266), leading to a MOTA (-1.2) and HOTA (-1.8) drop. {This demonstrates that the leftover detections in \cite{Zhang2022} are not reliable and insufficient for finding lost objects and it is therefore necessary to leverage a motion model providing better candidates}.
However, not every candidate is reliable, and relying solely on \textbf{+IoU} associations does not improve MOT performance (-1.1/-0.4 in MOTA/HOTA).
Adding visual cues with our \textbf{+Mixed} approach brings improvements, but the limited increase in MOTA (+0.1) evidences that this simple method still struggles to make correct assignments.
Differently, \textbf{\byteformer} considers visual and spatiotemporal information from the track, the candidate, and the context in a Transformer-based design, providing better decisions to prevent undetected tracks from being paused.

\noindent\textbf{Longer trajectories with \byteformer.}
{As illustrated in \cref{fig:ablation-savetracks}, the efficacy of \byteformer is evident in its ability to \emph{successfully} keep alive an extensive array of missing objects under different baselines.
We observe that most of those saved objects have low visibility (i.e., under heavy occlusions), proving that \byteformer is particularly good at mitigating instances where the detector exhibits a proclivity for failure. Accordingly, \byteformer correctly extends the resulting track trajectories \emph{in every tested tracker}, as demonstrated in \cref{fig:ablation-tracklength}.
}

\noindent\textbf{\byteformer component ablation.}
\byteformer relies on different components that ensure its proper operation and allow it to associate proposals and tracks accurately.
In \tablename~\ref{tab:ablation-byte-components}, we analyze the impact of the learned \texttt{[Miss.]} and \texttt{[Halluc.]} tokens, the spatiotemporal encoding, and the use of contextual information.

\byteformer may decide to pause a track either because it is a hallucinated track (\texttt{[Halluc.]} token), or because none of the candidates is suitable enough (\texttt{[Miss.]} token).
Relying solely on the \texttt{[Halluc.]} token (Line 2) yields negative results, resulting in an additional $+1501$ false positives compared to the baseline.
Conversely, if track termination is guided solely by the \texttt{[Miss.]} token (Line 3), the output remains marginally below the baseline with a decrease of $-0.1$ points in MOTA.
The integration of these two learned tokens leads to improved performance (Line 4) because taking into account both conditions for whether to associate a track more accurately represents real-world situations.

By adding our spatiotemporal encoding $\sts$ (Line 5), the MOTA score is further increased by $+0.2$ points.
Nonetheless, a high number of false negatives persist due to duplicated tracks occasionally kept alive.
These tracks negatively impact the system when kept active, and so far \byteformer has had no way of identifying them.
To address this issue, we integrate contextual proposals from nearby observations (Line 6), successfully reducing false negatives by $-733$ and resulting in a MOTA increase of $+0.4$ points.
The best results are achieved when all components are integrated into \byteformer (Line 7).

\begin{figure}[!t]
\centering
\begin{subfigure}{0.48\linewidth}
  \includegraphics[width=\textwidth]{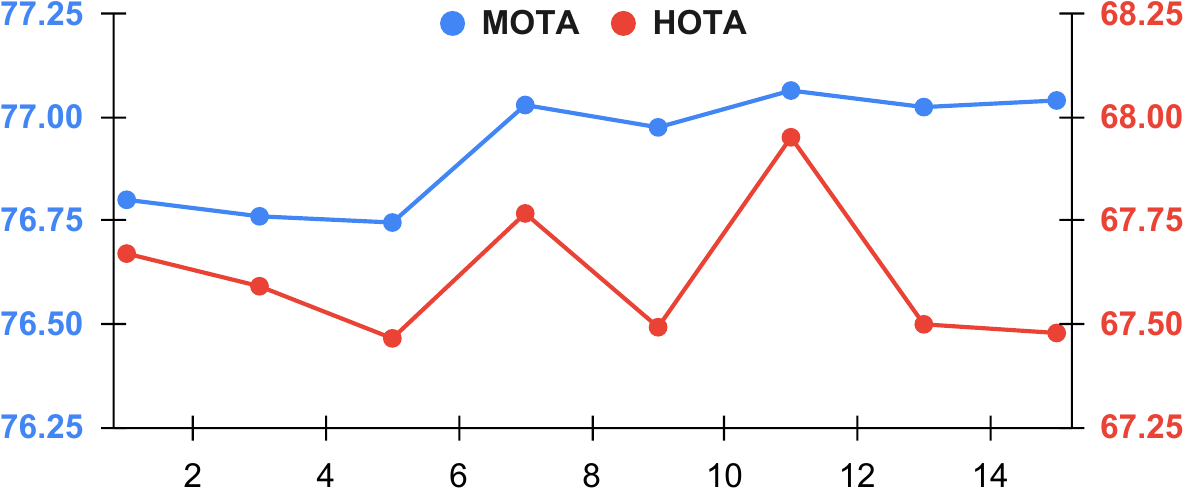}
  \caption{\textit{Track} length ($Z$)}
  \vspace{1mm}
\label{fig:ablation-seq_size}
\end{subfigure}
\begin{subfigure}{0.505\linewidth}
  \includegraphics[width=\textwidth]{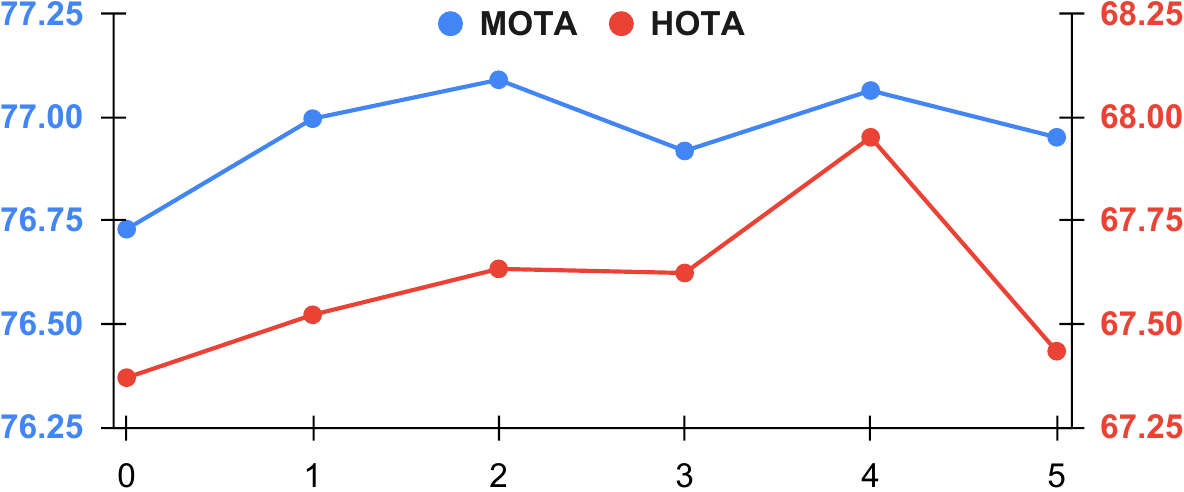}
  \caption{Number of \textit{contextual proposals} ($Q$)}
\label{fig:ablation-nei_size}
\end{subfigure}
\caption{Study of track length and number of contextual proposals used as input in our decision Transformer w.r.t. HOTA and MOTA performance.}
\end{figure}

\noindent\textbf{Track length, contextual proposal size.} {{Adhering to the definition of an online method,} \byteformer considers the past observations of a track and its interaction with neighboring objects, learning deep relationships between motion and appearance.}
On \cref{fig:ablation-seq_size}, we study the optimal amount of observations fed as input to \byteformer.
The HOTA curve contains noisier observations, whereas MOTA displays an upward trend that starts to converge at $Z = 11$ where HOTA also achieves the best score.
Regarding the maximum number of contextual proposals, from \cref{fig:ablation-nei_size}, we observe that both curves have a positive slope which decays when $Q > 4$. We hypothesize this is due to the additional contextual proposals being too distant and uninformative on the track's environment.

\begin{table*}[!t]
\setlength\extrarowheight{-3pt}
\centering
\caption{State-of-the-art comparison on MOT16, MOT17, and MOT20 test sets. $^{\star}$ means that the offline interpolation and the per-sequence thresholds in ByteTrack~\cite{Zhang2022} and OC-SORT~\cite{Cao2022_OCSORT} are removed for fair comparison.
$^{\dagger}$ and $^{\ddagger}$ indicate reproduced results for GHOST~\cite{Seidenschwarz2022} and StrongSORT~\cite{du2023strongsort} on MOT16 and for CenterTrack~\cite{Zhou2020b} on MOT20, respectively, due to their unavailability in the original works.
Private detections are used. \byteformer consistently improves all baseline trackers in almost every metric, as shown in \textbf{bold}. Best results are highlighted in blue.
}
\resizebox{0.98\linewidth}{!}{%
\begin{NiceTabular}{@{}l@{\hskip -5pt} c@{\hskip 1pt}l@{\hskip 1pt}c@{\hskip 1pt}l@{\hskip 3pt}c@{\hskip 1pt}l@{\hskip 2pt}c@{\hskip 1pt}c@{\hskip 4pt} | c@{\hskip 1pt}l@{\hskip 1pt}c@{\hskip 1pt}l@{\hskip 3pt}c@{\hskip 1pt}l@{\hskip 2pt}c@{\hskip 1pt}c@{\hskip 4pt} | c@{\hskip 1pt}l@{\hskip 1pt} c@{\hskip 1pt}l@{\hskip 3pt}c@{\hskip 1pt}l@{\hskip 2pt}c@{\hskip 1pt}c@{}}
\CodeBefore
   \rowcolors{30}{}{buscasota}
\Body
\hline \\[-6pt]
& & & \multicolumn{3}{c}{\textbf{MOT16}} & & & &  & \multicolumn{4}{c}{\textbf{MOT17}} & & & & &  \multicolumn{4}{c}{\textbf{MOT20}} \\

 & \textbf{MOTA}$\uparrow$ && \textbf{HOTA}$\uparrow$ && \textbf{IDF1}$\uparrow$ && \textbf{IDSW}$\downarrow$ && \textbf{MOTA}$\uparrow$ && \textbf{HOTA}$\uparrow$ && \textbf{IDF1}$\uparrow$ && \textbf{IDSW}$\downarrow$ && \textbf{MOTA}$\uparrow$ && \textbf{HOTA}$\uparrow$ && \textbf{IDF1}$\uparrow$ && \textbf{IDSW}$\downarrow$ & \\[-1pt]

\midrule
\midrule

TubeTK~\cite{Pang2020} & 66.9 && 50.8 && 62.2 && 1236 && 63.0 && 48.0 && 58.6 && 5727 && -- && -- && -- && -- & \\
CTracker~\cite{Peng2020} & 67.6 && 48.8 && 57.2 && 1897 && 66.6 && 49.0 && 57.4 && 5529 && -- && -- && -- && -- & \\
QDTrack~\cite{Pang2021} & 69.8 && 54.5 && 67.1 && 1097 && 68.7 && 53.9 && 66.3 && 3378 && -- && -- && -- && -- & \\
TraDeS~\cite{Wu2021} & 70.1 && 53.2 && 64.7 && 1144 && 69.1 && 52.7 && 63.9 && 3555 && -- && -- && -- && -- & \\
MTrack~\cite{Yu2022} & 72.9 && -- && 74.3 && 642 && 72.1 && -- && 73.5 && 2028 && 63.5 && -- && 69.2 && 6031 \\
MeMOT~\cite{Cai2022} & 72.6 && 57.4 && 69.7 && 845 && 72.5 && 56.9 && 69.0 && 2724 && 63.7 && 54.1 && 66.1 && 1938 \\
{MeMOTR~\cite{Gao_2023_ICCV}} & -- && -- && -- && -- && 72.8 && 58.8 && 71.5 && 1902 && -- && -- && -- && -- & \\ 
GSDT~\cite{Wang2021c} & 74.5 && 56.6 && 68.1 && 1229 && 73.2 && 55.2 && 66.5 && 3891 && 67.1 && 53.6 && 67.5 && 3230  \\

{Decode-MOT \cite{TBMotion}} & 74.7 && 60.2 &&73.0 && 1094 && 73.2 && 59.6 && 72.0 && 3363 && 67.2 && 54.5 && 69.0 && 2805 \\
MOTR~\cite{Zeng2022} & -- && -- && -- && -- && 73.4 && 57.8 && 68.6 && 2439 && -- && -- && -- && -- & \\
OUTrack~\cite{Liu2022} & 74.2 && 59.2 && 71.1 && 1328 && 73.5 && 58.7 && 70.2 && 4122 && 68.6 && 56.2 && 69.4 && 2223 \\
FairMOT~\cite{Zhang2021b} & 75.7 && 61.6 && 75.3 && \rankB{621} && 73.7 && 59.3 && 72.3 && 3303 && 61.8 && 54.6 && 67.3 && 5243 \\
{TrackFormer}~\cite{Meinhardt2021} & -- && -- && -- && -- && 74.1 && 57.3 && 68.0 && 2829 && 68.6 && 54.7 && 65.7 && 1532 \\
{TransTrack}~\cite{Sun2020b} & -- && -- && -- && -- && 74.5 && - && 63.9 && 3663 && 64.5 && -- && 59.2 && 3565 \\
AOH~\cite{Jiang2022} & -- && -- && -- && -- && 75.1 && 59.6 && 72.6 && 3312 && 67.9 && 55.1 && 70.0 && 2698 & \\ 
GTR~\cite{Zhou2022b} & -- && -- && -- && -- && 75.3 && 59.1 && 71.5 && 2859 && -- && -- && -- && -- & \\
CrowdTrack~\cite{Stadler2021} & -- && -- && -- && -- && 75.6 && 60.3 && 73.6 && 2544 && 70.7 && 55.0 && 68.2 && 3198 \\
OC-SORT$^{\star}$~\cite{Cao2022_OCSORT} & -- && -- && -- && -- && 76.0 && 61.7 && 76.2 && 2199 && 73.1 && 60.5 && 74.4 && 1307 \\
SGT~\cite{Hyun2023} & 76.8 && 61.2 && 73.5 && 1276 && 76.3 && 60.6 && 72.4 && 4578 && 72.8 && 56.9 && 70.5 && 2649 \\
CorrTracker~\cite{Wang2021b} & 76.6 && 61.0 && 74.3  && 1709 && 76.5 && 60.7 && 73.6  && 3369 && 65.2 && -- && 69.1  && 5183 \\
ReMOT~\cite{Yang2021} & 76.9 && 60.1 && 73.2 && 742 && 77.0 && 59.7 && 72.0 && 2853 && -- && -- && -- && -- & \\
Unicorn~\cite{Yan2022} & -- && -- && -- && -- && 77.2 && 61.7 && 75.5 && 5379 && -- && -- && -- && -- & \\
MTracker~\cite{Zhang2022b} & -- && -- && -- && -- && 77.3 && -- && 75.9 && 3255 && 66.3 && -- && 67.7 && 2715 \\
{MO3TR\smaller{-YOLOX}}~\cite{Zhu2021} & -- && -- && -- && -- && 77.6 && 60.3 && 72.9 && 2847 && 72.3 && 57.3 && 69.0 && 2200 \\
CountingMOT~\cite{Ren2022} & 77.6 && 62.0 && 75.2 && 1087 && 78.0 && 61.7 && 74.8 && 3453 && 70.2 && 57.0 && 72.4 && 2795 \\[-1pt]

\midrule
\midrule

CenterTrack$^{\ddagger}$~\cite{Zhou2020b} & 69.6 && -- && 60.7 && 2124 && 67.8 && 52.2 && 64.7 && 3039 && 45.8 && 31.8 && 36.6 && 6296 \\
\makecell{\plusours{} \textbf{(ours)} \vspace*{-1mm}\\ \ } & \makecell{\textbf{70.4} \vspace*{-1mm}\\ \gooddelta{+0.8}} && \makecell{\textbf{55.7} \vspace*{-1mm}\\ \baddelta{-}} && \makecell{\textbf{69.7} \vspace*{-1mm}\\ \gooddelta{+9.0}} && \makecell{\textbf{927} \vspace*{-1mm}\\ \gooddelta{\text{-}1197}} && \makecell{\textbf{68.9} \vspace*{-1mm}\\ \gooddelta{+1.1}} && \makecell{\textbf{55.1} \vspace*{-1mm}\\ \gooddelta{+2.9}} && \makecell{\textbf{68.8} \vspace*{-1mm}\\ \gooddelta{+4.1}} && \makecell{\textbf{2817} \vspace*{-1mm}\\ \gooddelta{\text{-}222}} && \makecell{\textbf{49.5} \vspace*{-1mm}\\ \gooddelta{+3.7}} && \makecell{\textbf{44.2} \vspace*{-1mm}\\ \gooddelta{+12}} && \makecell{\textbf{58.0} \vspace*{-1mm}\\ \gooddelta{+21}} && \makecell{\textbf{1370} \vspace*{-1mm}\\ \gooddelta{\text{-}4926}} \\[-2pt]

\midrule

TransCenter~\cite{Xu2023} & 75.7 && 56.9 && 65.9 && 1717 && 76.2 && 56.6 && 65.5 && 5427 && 72.9 && 50.2 && 57.7 && 2625 \\
\makecell{\plusours{} \textbf{(ours)} \vspace*{-1mm}\\ \ } & \makecell{\textbf{75.7} \vspace*{-1mm}\\ \baddelta{+0.0}} && \makecell{\textbf{61.9} \vspace*{-1mm}\\ \gooddelta{+5.0}} && \makecell{\textbf{74.5} \vspace*{-1mm}\\ \gooddelta{+8.6}} && \makecell{\textbf{1038} \vspace*{-1mm}\\ \gooddelta{\text{-}679}} && \makecell{\textbf{76.2} \vspace*{-1mm}\\ \baddelta{+0.0}} && \makecell{\textbf{61.7} \vspace*{-1mm}\\ \gooddelta{+5.1}} && \makecell{\textbf{74.1} \vspace*{-1mm}\\ \gooddelta{+8.6}} && \makecell{\textbf{3282} \vspace*{-1mm}\\ \gooddelta{\text{-}2145}} && \makecell{\rankC{\textbf{73.9}} \vspace*{-1mm}\\ \gooddelta{+1.0}} && \makecell{\textbf{58.8} \vspace*{-1mm}\\ \gooddelta{+8.6}} && \makecell{\textbf{72.4} \vspace*{-1mm}\\ \gooddelta{+15}} && \makecell{\textbf{1756} \vspace*{-1mm}\\ \gooddelta{\text{-}869}} \\[-2pt]

\midrule

GHOST$^{\dagger}$~\cite{Seidenschwarz2022} & \rankB{78.3} && {63.0} && \rankC{77.4} && \rankC{709} && 78.7 && \rankC{62.8} && \rankB{77.1} && \rankC{2325} && 73.7 && \rankB{61.2} && 75.2 && 1264 \\
\makecell{\plusours{} \textbf{(ours)} \vspace*{-1mm}\\ \ } & \makecell{\textcolor{blue}{\textbf{78.5}} \vspace*{-1mm}\\ \gooddelta{+0.2}} && \makecell{\rankB{\textbf{63.2}} \vspace*{-1mm}\\ \gooddelta{+0.2}} && \makecell{\rankB{\textbf{77.5}} \vspace*{-1mm}\\ \gooddelta{+0.1}} && \makecell{\rankB{\textbf{707}} \vspace*{-1mm}\\ \gooddelta{\text{-}2}} && \makecell{{\textbf{79.0}} \vspace*{-1mm}\\ \gooddelta{+0.3}} && \makecell{\rankB{\textbf{62.9}} \vspace*{-1mm}\\ \gooddelta{+0.1}} && \makecell{\rankC{77.0} \vspace*{-1mm}\\ \baddelta{\text{-}0.1}} && \makecell{\rankB{\textbf{2295}} \vspace*{-1mm}\\ \gooddelta{\text{-}30}} && \makecell{\rankB{\textbf{74.2}} \vspace*{-1mm}\\ \gooddelta{+0.5}} && \makecell{\textbf{61.3} \vspace*{-1mm}\\ \gooddelta{+0.1}} && \makecell{\rankB{75.1} \vspace*{-1mm}\\ \baddelta{\text{-}0.1}} && \makecell{\rankC{\textbf{1251}} \vspace*{-1mm}\\ \gooddelta{\text{-}13}} \\[-2pt]

\midrule

StrongSORT$^{\dagger}$~\cite{du2023strongsort} & \rankB{78.3} && {63.8} && \rankC{78.9} && {437} && 78.3 && \rankB{63.5} && \rankB{78.5} && \rankB{1446} && 72.2 && 61.5 && 75.9 && 1066 \\
\makecell{\plusours{} \textbf{(ours)} \vspace*{-1mm}\\ \ } & \makecell{{\textbf{78.4}} \vspace*{-1mm}\\ \gooddelta{+0.1}} && \makecell{\textcolor{blue}{\textbf{64.2}} \vspace*{-1mm}\\ \gooddelta{+0.4}} && \makecell{\textcolor{blue}{\textbf{79.5}} \vspace*{-1mm}\\ \gooddelta{+0.6}} && \makecell{\textcolor{blue}{\textbf{434}} \vspace*{-1mm}\\ \gooddelta{\text{-}3}} && \makecell{\rankB{\textbf{78.6}} \vspace*{-1mm}\\ \gooddelta{+0.3}} && \makecell{\textcolor{blue}{\textbf{63.9}} \vspace*{-1mm}\\ \gooddelta{+0.4}} && \makecell{\textcolor{blue}{\textbf{79.2}} \vspace*{-1mm}\\ \gooddelta{+0.7}} && \makecell{\textcolor{blue}{\textbf{1428}} \vspace*{-1mm}\\ \gooddelta{\text{-}18}} && \makecell{\rankB{\textbf{72.7}} \vspace*{-1mm}\\ \gooddelta{+0.5}} && \makecell{\textcolor{blue}{\textbf{61.8}} \vspace*{-1mm}\\ \gooddelta{+0.3}} && \makecell{\textcolor{blue}{\textbf{76.3}} \vspace*{-1mm}\\ \gooddelta{+0.4}} && \makecell{\rankC{\textbf{1006}} \vspace*{-1mm}\\ \gooddelta{\text{-}60}} \\[-2pt]

\midrule

ByteTrack$^{\star}$~\cite{Zhang2022} & \rankB{78.2} && 62.8 && 77.2 && 892 && \rankC{78.9} && \rankC{62.8} && \rankB{77.1} && 2363 && \rankB{74.2} && 60.4 && \rankC{74.5} && \rankB{925} \\ 
\makecell{\plusours{} \textbf{(ours)} \vspace*{-1mm}\\ \ } & \makecell{\textcolor{blue}{\textbf{78.5}} \vspace*{-1mm}\\ \gooddelta{+0.3}} && \makecell{\rankB{\textbf{63.3}} \vspace*{-1mm}\\ \gooddelta{+0.5}} && \makecell{\rankB{\textbf{77.9}} \vspace*{-1mm}\\ \gooddelta{+0.7}} && \makecell{\textbf{743} \vspace*{-1mm}\\ \gooddelta{\text{-}145}} && \makecell{\textcolor{blue}{\textbf{79.3}} \vspace*{-1mm}\\ \gooddelta{+0.4}} && \makecell{\rankC{\textbf{63.1}} \vspace*{-1mm}\\ \gooddelta{+0.3}} && \makecell{\rankC{\textbf{77.7}} \vspace*{-1mm}\\ \gooddelta{+0.6}} && \makecell{\rankC{\textbf{2358}} \vspace*{-1mm}\\ \gooddelta{\text{-}5}} && \makecell{\textcolor{blue}{\textbf{74.5}} \vspace*{-1mm}\\ \gooddelta{+0.3}} && \makecell{\rankC{\textbf{60.5}} \vspace*{-1mm}\\ \gooddelta{+0.1}} && \makecell{74.4 \vspace*{-1mm}\\ \baddelta{\text{-}0.1}} && \makecell{\textcolor{blue}{\textbf{920}} \vspace*{-1mm}\\ \gooddelta{\text{-}5}} \\[-2pt]
\midrule

\end{NiceTabular}}

\label{tab:sota-byte}
\end{table*}

\subsection{State-of-the-Art Comparisons} \label{subsec:sota}
By design, \byteformer can be seamlessly incorporated into any existing online TbD tracker. {To illustrate its performance, we extensively integrate \byteformer into five diverse state-of-the-art trackers and compare them against the current state-of-the-art in online MOT.
Our base trackers include the center-based CenterTrack~\cite{Zhou2022} (CNN network) and TransCenter~\cite{Xu2023} (Transformer network); as well as the YOLOX-based ByteTrack~\cite{Zhang2022} (IoU matching), StrongSORT~\cite{du2023strongsort} (appearance-enhanced association), and GHOST~\cite{Seidenschwarz2022} (attentive Re-ID scheme).}
Evaluations were conducted on the test sets of MOT16~\cite{Milan2016a}, MOT17~\cite{Milan2016a}, and MOT20 \cite{Dendorfer2020}.
As shown in \cref{tab:sota-byte}, \emph{\byteformer consistently improves the performance of all trackers in every benchmark for nearly all metrics, without requiring training on any real MOT data nor necessitating to be fine-tuned for any tracker}.

Remarkably, \byteformer \emph{drastically} enhances both CenterTrack and TransCenter without the necessity for a recent state-of-the-art detector.
For instance, in CenterTrack, we achieve a boost of $+12$ HOTA and $+21$ IDF1 in MOT20.
Similarly, TransCenter also gets significantly improved due to a marked reduction in IDSW, thereby bolstering HOTA (e.g., $+5.1$/$+8.6$ in MOT17/20) and IDF1 (e.g., $+8.6$/$+15$ in MOT17/20).
When paired with high-performing trackers such as ByteTrack and StrongSORT that rely on a potent YOLOX detector~\cite{Ge2021}, \byteformer sets a new state-of-the-art for online multi-object tracking.
{Furthermore, \byteformer can also join efforts with identity-preserving methods like the advanced Re-ID mechanism in GHOST~\cite{Seidenschwarz2022} to further enhance its performance.}

Lastly, recent tracking-by-attention methods~\cite{Meinhardt2021, Zhu2021, Zeng2022, Zhou2022b, Gao_2023_ICCV} strive to create a fully end-to-end architecture that performs both object detection and track-detection matching within a single network.
However, this streamlined process hinders their ability to easily incorporate new elements, such as a more powerful detector.
This is illustrated by MOT3TR-YOLOX~\cite{Zhu2021}, a recent model that, despite adopting a more powerful YOLOX detection backbone, still underperforms TransCenter+\byteformer by $-1.4$ HOTA, $-1.2$ IDF1 in MOT17 and by $-1.5$ HOTA $-3.4$ IDF1 in MOT20.
This underscores the superior performance of TbD methods and the opportunities that \byteformer brings, offering a plug-and-play module that systematically enhances state-of-the-art TbD trackers in a fully online manner without the need for retraining.

\begin{figure*}[!t]
\centering
  \subfloat[]{\includegraphics[width=0.24\linewidth]{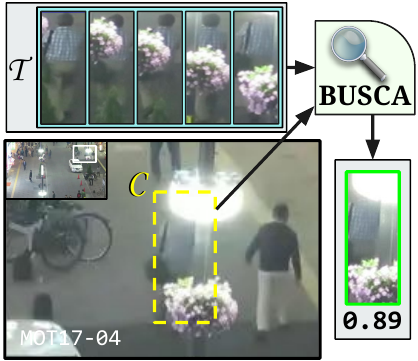}
  \label{fig:qualitative-01}}
\hfil
  \subfloat[]{\includegraphics[width=0.24\linewidth]{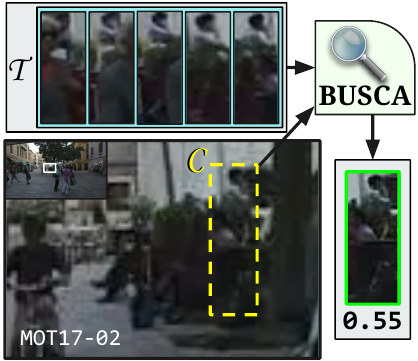}
  \label{fig:qualitative-02}}
  \subfloat[]{\includegraphics[width=0.24\linewidth]{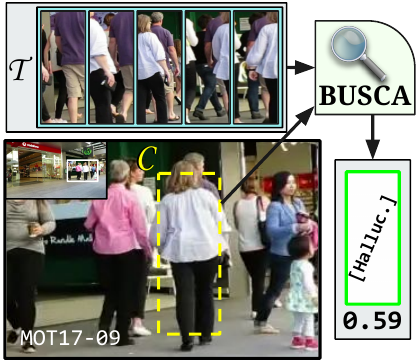}
  \label{fig:qualitative-03}}
\hfil
  \subfloat[]{\includegraphics[width=0.24\linewidth]{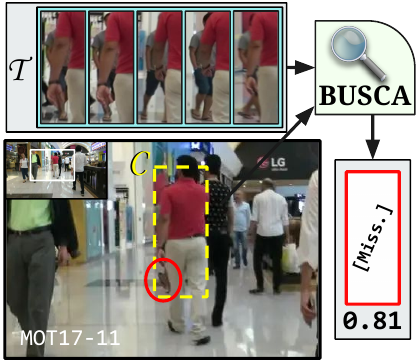}
  \label{fig:qualitative-04}}

\caption{Qualitative examples of \byteformer integrated into ByteTrack~\cite{Zhang2022} for MOT17-val~\cite{Dendorfer2021}.
\protect\subref{fig:qualitative-01}, \protect\subref{fig:qualitative-02}, and \protect\subref{fig:qualitative-03} depict correct predictions while \protect\subref{fig:qualitative-04} shows a scenario where \byteformer incorrectly labels the pedestrian wearing a gray shirt as `missing', even though the individual's left foot (highlighted with a red circle) remains visible. The values indicate the assignment confidence.}
\label{fig:qualitative}
\end{figure*}

\subsection{Qualitative Results} \label{subsec:quali}
\cref{fig:qualitative} showcases a series of qualitative visualizations.
In \cref{fig:qualitative-01}, the YOLOX detector~\cite{Ge2021} fails to detect the person obscured by the street lamp and flowers due to substantial occlusion.
However, with \byteformer, we can successfully preserve his identity.
A similar scenario unfolds in \cref{fig:qualitative-02}, where the pedestrian in the background is accurately identified by \byteformer despite his minimal size and the scarce visibility of only his head.
\cref{fig:qualitative-03} illustrates a clearly spurious track created by ByteTrack~\cite{Zhang2022} that does not correlate to any specific person.
\byteformer correctly identifies it as a hallucination and deactivates it, effectively preventing any further false positives.
Lastly, in \cref{fig:qualitative-04}, due to the noisy track and the almost total occlusion, the pedestrian wearing a gray shirt is incorrectly labeled as missing, even though his left foot can still be spotted behind the man in red.
Additional videos are provided in the supplementary material.

\section{Conclusion\label{sec:byteformer-conclusions}}
{In this work, we present \byteformer, an innovative and plug-and-play framework that can enhance any online tracking-by-detection system to persistently track undetected objects in a fully online fashion. This {implies} that \byteformer \textit{does not} alter the outputs of previous time steps or access future frames.  To achieve this, our novel Decision Transformer associates tracks with proposals having both visual and spatiotemporal information, maintaining the identity of tracks in a lightweight manner and without any need for fine-tuning.

We extensively validate our proposed method with five distinct trackers, bringing systematic performance improvements and setting new state-of-the-art results across different benchmarks.
{For future work, we aim to factor in extreme motions via nonlinear multi-candidate proposals, incorporate 3D multimodal cues, and explore the use of \byteformer to override previous tracking decisions and fix incorrect associations.
We hope that \byteformer can inspire future research towards fully online trackers without overly relying on the detectors.}

%
%
\newpage
\section*{Acknowledgements}
This work was partially supported by the EU ISFP PRECRISIS (ISFP-2022-TFI-AG-PROTECT-02-101100539) project, the EU WIDERA PATTERN (HORI-ZON-WIDERA-2023-ACCESS-04-01-101159751) project, MIAI@Grenoble Alpes (ANR-19-P3IA-0003), and the Spanish Ministerio de Ciencia e Innovación (grant numbers PID2020-112623GB-I00 and PID2021-128009OB-C32).
We thank Eloi Zablocki from Valeo.ai for the meaningful discussion.

\bibliographystyle{splncs04}
\bibliography{main}

\clearpage
\appendix

\vspace*{2mm}
\section*{\texorpdfstring{\centering \larger}{} Supplementary Material}
\vspace*{10mm}

\begin{figure*}[ht]
\centering
\includegraphics[width=0.95\linewidth]{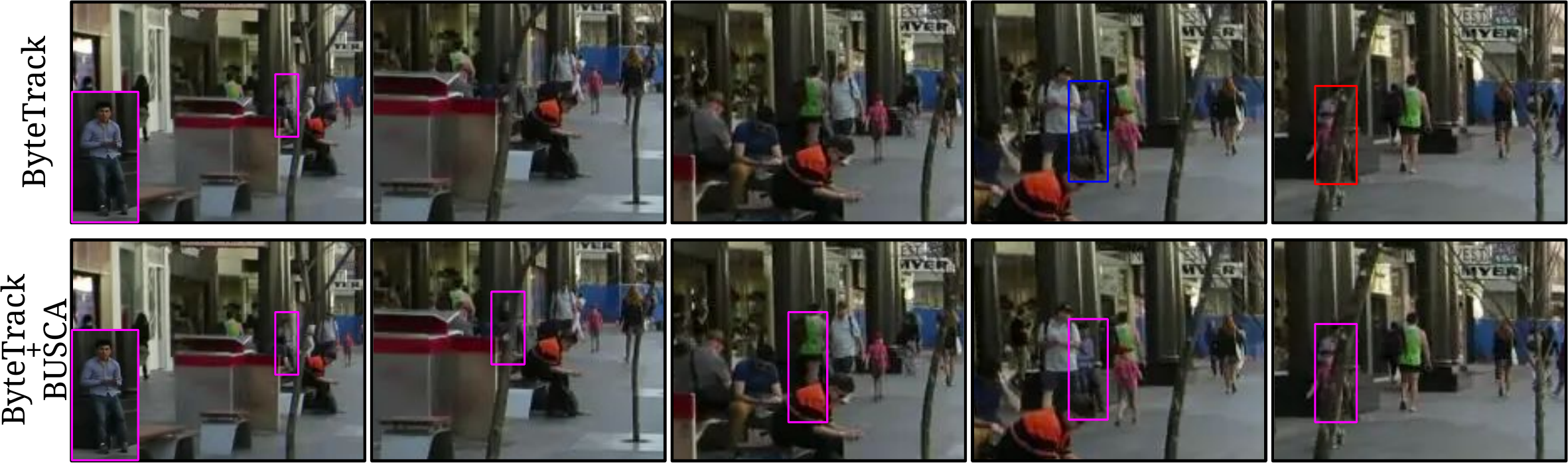}\vspace*{3mm}\\
\includegraphics[width=0.95\linewidth]{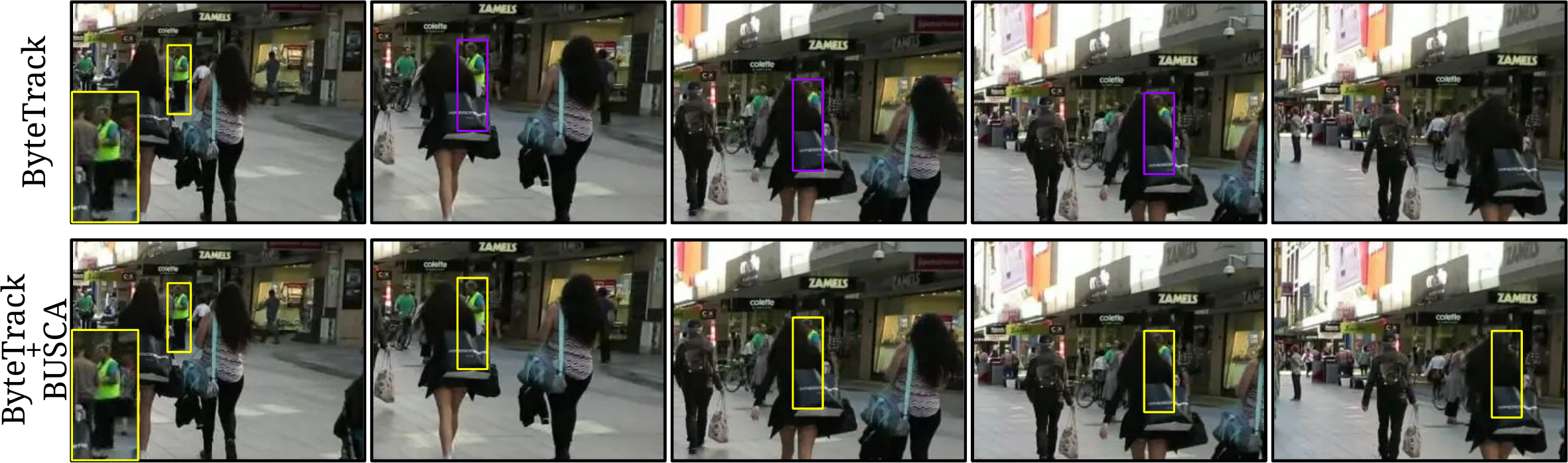}
\caption{Additional qualitative results showing the benefit of using \byteformer on different online TbD trackers like ByteTrack~\cite{Zhang2022}. We can see that \byteformer improves the trajectory consistency and continuity of the baseline trackers. Colors represent object identities.
Results are shown for only one subject to ease the visualization.
}
\label{fig:teaser}
\end{figure*}

\section{Introduction}
In this supplementary material, we show additional qualitative results in \cref{sec:quali_supp}, which demonstrate the benefits of using \byteformer.
Following this, we underline the fact that \byteformer is a generic framework (\cref{sec:generality}) applicable to any online TbD method by definition, which is backed up by the experimental results in the main paper.
We then provide implementation details about the network architecture, training, and inference parameters in \cref{sec:detial}, and discuss in \cref{sec:implementation-multimodal} the characteristics of the naive baselines introduced in the main paper.
Subsequently, we further explain in \cref{sec:neibhood} how to calculate the vicinity of a track and select its neighbors as contextual proposals and further discuss the impact of \byteformer on object trajectories in \cref{sec:track_traj_supp}.
Then, we detail how to encode spatiotemporal information for the learned tokens in \cref{sec:sp_ltks} and demonstrate the effect of the \texttt{[SEP]} token in \byteformer's performance in \cref{sec:ablation_sep}.

BUSCA generalizes well in different trackers and scenarios without being trained on real MOT datasets. Nonetheless, we discuss in \cref{sec:finetuning_busca} the possible benefits of training and fine-tuning \byteformer on real in-domain data.
Additionally, in \cref{sec:dancetrack} we show how our Kalman-based motion model handles complex motions like the dancing scenes found in DanceTrack~\cite{sun2022dance}, and in \cref{sec:bdd100k} we demonstrate the performance of \byteformer for categories different than humans.

Trackers like ByteTrack \cite{Zhang2022} and StrongSORT \cite{du2023strongsort} employ \emph{offline} interpolation methods to further improve their MOT performance by modifying past predictions with future frame information. Differently, \byteformer strictly respects the \emph{online} definition.
Although they are not directly comparable, to give an idea of the performance differences, we show in \cref{sec:vsoffline} the offline version of ByteTrack and StrongSORT versus their fully online versions with \byteformer, and the potential benefits of embedding \byteformer in an offline tracker.
Lastly, to facilitate the analysis of \byteformer's performance, we include its sequence-wise results in \cref{sec:detailed}.

\section{Additional Qualitative Results \label{sec:quali_supp}}
The efficacy of online tracking by detection (TbD) is largely dependent on the underlying detectors.
Issues such as object loss or identity switches often arise when these detectors miss some of the objects in the scene.
In order to showcase how incorporating \byteformer into a TbD system can enhance track consistency and continuity, we provide several illustrative examples.
More specifically, \cref{fig:teaser} elucidates the advantages of integrating \byteformer into ByteTrack~\cite{Zhang2022}, while Figs. \ref{fig:quali_strongsort}, \ref{fig:quali_ghost}, \ref{fig:quali_transcenter}, and \ref{fig:quali_centertrack} (found at the end of this supplementary material) visually demonstrate the application of \byteformer to StrongSORT~\cite{du2023strongsort}, GHOST~\cite{Seidenschwarz2022}, TransCenter~\cite{Xu2023}, and CenterTrack~\cite{Zhou2020b}, respectively.
The visualizations display the results for a single subject for ease of visualization.
To view the results for every object in the sequences, the reader is referred to the videos included in this supplementary material.

\section{On the Generality of \byteformer} \label{sec:generality}
We claim that \byteformer's design allows it to be applied to \textit{any} online TbD tracker.
This is because, by definition~\cite{Dai2022}, online TbD trackers (i) detect objects in the current frame and (ii) link them to existing tracks.
During this process, it is natural for some detections to remain unmatched, leading either to the creation of new tracks or their dismissal.
Similarly, it is also common for some tracks to be unmatched and, therefore, paused.
\byteformer introduces an additional step that (iii) propagates these unmatched tracks without requiring additional detections, a feature applicable to the results of any online TbD track assignment procedure.
Hence, we assert the adaptability of \byteformer \textit{any} online TbD tracker.

\section{Network, Inference and Training Details \label{sec:detial}}

\textbf{Network architecture.}
\byteformer's decision Transformer is composed of $L = 4$ encoder blocks, each one of them comprising a $4$-headed multi-attention layer and a $1024$-size feed-forward layer.
The internal dimension of the Transformer is $D^{Tr} = 512$ channels.
For our spatiotemporal encoding, the scaling factors are set as $\sigma^t = 2$, $\sigma^s = 15$, and $\sigma^d = 15$.

The extraction of the appearance features $a$ of each observation is performed with a ResNet-50~\cite{He2016} with an extra fully-connected layer for downsampling and the domain adaptation mechanism described in \cite{Seidenschwarz2022} (i.e., samples are normalized using the mean and variance of the batch, instead of the learned ones).
To this end, the coordinates $c$ of each observation are cropped to $128 \times 384$~px and fed to the feature extractor, yielding a $512$-channel embedding.
This appearance model is pre-trained on the Re-ID dataset Market-1501~\cite{Zheng2015} and \textit{is not trained} on any MOT data.

\textbf{Inference parameters.}
Following the experimental results discussed in Sec. \textcolor{red}{5.2} of the main paper, candidate $\mathcal{B}$ is generated using a simple-yet-effective Kalman filter~\cite{Kalman1961} that forecasts a new observation for $\tau$ at the present frame, while the contextual proposal set $\mathcal{C}$ comprises the $Q=4$ closest observations within the neighborhood of $\tau$.
If $\mathcal{B}$ is chosen as the correct candidate, we will keep $\tau$ active and update it with the new Kalman-based observation.
Otherwise, or if $\tau$ has fewer than $Z=11$ observations (indicating low reliability in the Decision Transformer's prediction), we will not update the track and let the underlying base tracker handle it through its usual process (either deactivating the track or increasing its inactive counter).

\textbf{Training parameters.}
\byteformer parameters are randomly initialized and trained via label-smoothed cross-entropy loss for $25$ epochs using an AdamW optimizer~\cite{Loshchilov2019} with a dropout regularization probability of $0.1$ and a batch size of $256$.
We set the weight decay at $1 \times 10^{-5}$, with the initial learning rate established at $2 \times 10^{-5}$.
Following the 20th epoch, we reduce the learning rate by a factor of $10$.

For each training sample, we randomly choose an object identity from the dataset and construct a track $\tau$ by sampling $Z=11$ observations, ensuring a maximum separation of 10 frames between consecutive observations.
Subsequently, we randomly pick a frame within a range of 20 frames from the last observation and select 5 objects near $\tau$ to form a proposal set.
From these proposals, we designate the ground truth annotation of $\tau$ as the positive candidate, while objects with an overlap smaller than $0.5$ are marked as negatives (others are ignored).
Additionally, we set a $15\%$ probability of not sampling any positives, in which case the \texttt{[Miss.]} token will be considered the correct option within the proposal set, and a $1\%$ chance of randomly eliminating some proposals.
Furthermore, observations within $\tau$ are subject to alteration with a $1\%$ probability, either through removal or replacement with different object observations.
If at least $5\%$ of the observations in $\tau$ correspond to a different object, the track is deemed unreliable, and the \texttt{[Halluc.]} token is considered the correct option.

The entire training process took roughly $28$ hours, utilizing a single NVIDIA Quadro RTX 8000.
As highlighted in the main paper, \byteformer does not necessitate training aimed at any particular tracker or real MOT dataset. {\byteformer is trained on a subset of 100 MOTSynth~\cite{Fabbri2021} videos, and does not need to be fine-tuned.}
Therefore, we consistently use \emph{the same weights} across all experiments featured in our main paper.

\section{Implementation of Naive Approaches\label{sec:implementation-multimodal}}
In Sec. \textcolor{red}{5.2} of the main paper, we present the results of approaches proposed by us that follow the same philosophy as \byteformer (i.e., handling those tracks without matching detections in an online manner).
Specifically, the IoU-based approach computes the intersection over union between the proposals generated by \byteformer and the object's last known bounding box, assigning as correct proposal to the one with the highest overlap.
On the other hand, the Mixed method uses overlap as a threshold to filter out the unrealistic proposals ($\text{IoU} < 0.7$) and, in a second step, it uses the cosine similarity between the ReID features (extracted with GHOST~\cite{Seidenschwarz2022}) of proposals and the last observation of the track to determine the most suitable match.

\section{Neighborhood Computation \label{sec:neibhood}}
The intention behind our contextual proposals $\mathcal{C}$ is to equip \byteformer with a broader understanding of the scene.
To this end, as stated in the main paper, we pool the $Q$ closest observations within the neighborhood of track $\tau$, $V(\tau)$.
We envision $V(\tau)$ possessing two characteristics: first, observations that are spatially adjacent to $\tau$ are deemed closer neighbors, and second, a clear demarcation is maintained between foreground and background objects.
To meet these ends, we calculate the distance $\phi(\tau, o)$ between the last known coordinates of track $\tau$ and an observation $o$ as follows:
\begin{equation}
\phi(\tau, o) =  \text{Eucl}(\tau, o) * \text{Ratio}(\tau, o)
\end{equation}
\begin{equation}
\text{Eucl}(\tau, o) = \sqrt{(x_\tau - x_o)^2 + (y_\tau - y_o)^2}
\end{equation}
\begin{equation}
\text{Ratio}(\tau, o) = \max{\left(\frac{\sqrt{w_\tau * h_\tau}}{\sqrt{w_o * h_o}}, \frac{\sqrt{w_o * h_o}}{\sqrt{w_\tau * h_\tau}}\right)}
\end{equation}
Thus, in defining this distance, we consider not only the Euclidean distance between centers $\text{Eucl}(\cdot, \cdot)$, but also penalize this value based on the difference in object sizes $\text{Ratio}(\cdot, \cdot)$. 
This approach is driven by the fact that object size serves as a strong indicator of depth in the objects within a scene~\cite{Nasseri2021}.

Consequently, the neighborhood of $\tau$ can be defined as $V(\tau) = \{ o \in \mathcal{T} \setminus\mathcal{T}_u\} \mid \phi(\tau, o) < \nu_\tau \}$, where $\nu_\tau$ represents the maximum distance within which an observation is considered a neighbor.
The maximum distance acts as a variable parameter, fluctuating about the area of the track as discussed in~\cite{Bertinetto2016}.
This can be computed using the following equation:
\begin{equation}
\nu_{\tau} = \sqrt{\left(w_\tau + \zeta\left(w_\tau + h_\tau\right)\right) * \left(h_\tau + \zeta\left(w_\tau + h_\tau\right)\right)}
\end{equation}
Here, $\zeta=1$ is employed as a scaling factor that governs the growth of $\nu$ concerning $\tau$.

\section{Impact of \byteformer on Object Trajectories \label{sec:track_traj_supp}}
\byteformer can be incorporated into any online tracking-by-detection system, enhancing its capabilities to persistently track those objects missed by the detector.
As can be seen in Fig.~\textcolor{red}{3} of the main paper, \byteformer primarily focuses on those objects where the detector most frequently fails, specifically those with minimal visibility.
This has the added advantage of extending the average lifespan of the tracks, thereby enhancing their trajectory consistency and continuity.

To conduct the experiment shown in Fig. \textcolor{red}{3a}, we used the visibility attributes contained within the MOT17~\cite{Dendorfer2021} ground truth.
Thus, for every object that \byteformer finds and that would otherwise have been paused by the tracker, we refer to its corresponding annotation and visibility attribute.
This is performed for each combination of tracker+\byteformer studied, confirming that \byteformer is capable of identifying a substantial number of objects with extremely low visibility, due to occlusions.

About the effect of \byteformer on the track length, Fig.~\textcolor{red}{3b} illustrates the difference between using various standalone trackers and combining them with \byteformer.
For this experiment, we evaluated the results produced by each combination, quantifying the frequency of each ID reported (i.e., the length of each track).
As demonstrated, incorporating \byteformer engenders positive effects, amplifying both the median and average length of tracks for all five tested trackers.

\section{Learned Tokens \label{sec:sp_ltks}}
The appearance features $a$ of the learned proposals $\mathcal{L} = \{\text{\texttt{[Halluc.]}}, \text{\texttt{[Miss.]}}\}$, along with the separator token \texttt{[SEP]}, are initialized using a random Gaussian distribution and are trained end-to-end alongside the rest of the architecture.
There is no need for these features to pass through the appearance extractor, being directly fed to \byteformer's decision Transformer.

Regarding the coordinates component $c$ of the learned tokens, \texttt{[Miss.]} is given the same coordinates as the last known observation of $\tau$ and \texttt{[Halluc.]}, is computed by maximizing its distance w.r.t. $\tau$ in the spatiotemporal representation space (main paper, Sec. \textcolor{red}{4.3}).
Lastly, \texttt{[SEP]} tokens are given the same coordinates as the proposals they delimit.

\begin{table}[t!]
\centering
\caption{Ablation of the effect of the \texttt{[SEP]} token on MOT17~\cite{Dendorfer2021} validation data.
\byteformer uses several \texttt{[SEP]} tokens to delimit every proposal $p \in \mathcal{P}$, for a total of $|\mathcal{P}|$ separator tokens.
In $\times 1$~\texttt{[SEP]}, we utilize a single token to delimit $\tau$ from $\mathcal{P}$.
ByteTrack~\cite{Zhang2022} is used as base tracker without its offline interpolation and per-sequence threshold, noted with $^{\star}$.}
\label{tab:ablation-sep}
\resizebox{0.70\linewidth}{!}{%
\begin{tabular}{@{}l|c@{\hskip 3pt}cc@{\hskip 3pt}cc@{\hskip 3pt}cc@{\hskip 3pt}c}
\toprule

& \multicolumn{2}{c}{\textbf{MOTA} $\uparrow$} & \multicolumn{2}{c}{\textbf{HOTA} $\uparrow$} & \multicolumn{2}{c}{\textbf{IDF1} $\uparrow$} & \multicolumn{2}{c}{\textbf{IDSW} $\downarrow$} \\
\hline
\hline

ByteTrack$^{\star}$  & 76.5 &  & 67.4 & & 79.4 &  & 165 &  \\
\plusours{} w/ $\times 1$~\texttt{[SEP]} & 76.8 &  \gooddelta{+0.3} & 67.3 & \baddelta{\text{-}0.1} & 78.8  & \baddelta{\text{-}0.8} & \rankA{162} & \gooddelta{\text{-}3}  \\
\plusours{} w/ $\times |\mathcal{P}|$~\texttt{[SEP]} & \rankA{77.1} &  \gooddelta{+0.6} & \rankA{67.6} & \gooddelta{+0.2} & \rankA{79.5} & \gooddelta{+0.1} & 166 & \baddelta{+1}  \\[-2pt]
\bottomrule
\end{tabular}}
\end{table}

\begin{table}[t!]
\centering
\caption{Ablation of training \byteformer on in-domain data for MOT17~\cite{Dendorfer2021}.
We test both training from scratch and fine-tuning \byteformer after training it on MOTSynth~\cite{Fabbri2021}.
The difference with the baseline is depicted next to each metric. ByteTrack \cite{Zhang2022} is used as base tracker without its offline interpolation and per-sequence threshold, noted with $^{\star}$.}
\label{tab:ablation-finetune}
\resizebox{0.80\linewidth}{!}{%
\begin{tabular}{@{}ll|c@{\hskip 3pt}cc@{\hskip 3pt}cc@{\hskip 3pt}cc@{\hskip 3pt}c}
\toprule

&& \multicolumn{2}{c}{\textbf{MOTA} $\uparrow$} & \multicolumn{2}{c}{\textbf{HOTA} $\uparrow$} & \multicolumn{2}{c}{\textbf{IDF1} $\uparrow$} & \multicolumn{2}{c}{\textbf{IDSW} $\downarrow$} \\
\hline
\hline

\parbox[t]{1mm}{\multirow{4}{*}{\rotatebox[origin=c]{90}{Val.}}} &ByteTrack$^{\star}$  & 76.5 &  & 67.4 & & 79.4 &  & 165 &  \\ 
& \plusours{} \textbf{(MOT17 train)} & 76.8 &  \gooddelta{+0.3} & 67.4 & \baddelta{+0.0} & 79.1  & \baddelta{\text{-}0.3} & 167 & \baddelta{+2} \\
& \plusours{} \textbf{(MOTSynth train)}  & 77.1 &  \gooddelta{+0.6} & 67.6 & \gooddelta{+0.2} & 79.5 & \gooddelta{+0.1} & 166 & \baddelta{+1}  \\
& \plusours{} \textbf{(MOT17 fine-tune)} & \rankA{77.2} &  \gooddelta{+0.7} & \rankA{67.9} & \gooddelta{+0.5} & \textbf{79.8}  & \gooddelta{+0.4} & \textbf{150} & \gooddelta{\text{-}15}  \\

\midrule
\midrule
\parbox[t]{3mm}{\multirow{3}{*}{\rotatebox[origin=c]{90}{Test}}} &ByteTrack$^{\star}$  & 78.9 &  & 62.8 && \rankB{77.1} && 2363 &  \\
& \plusours{} \textbf{(MOT17 train)}  & \rankA{79.3} &  \gooddelta{+0.4} & \rankA{63.1} & \gooddelta{+0.3} & \rankC{77.7} & \gooddelta{+0.6} & \rankC{2358} & \gooddelta{\text{-}5} \\
& \plusours{} \textbf{(MOT17 fine-tune)} & \rankA{79.3} &  \gooddelta{+0.4} & \rankA{63.1} & \gooddelta{+0.3} & \textbf{78.8} & \gooddelta{+0.9} & \textbf{2349} & \gooddelta{\text{-}14}  \\
\bottomrule
\end{tabular}}
\end{table}

\section{Effect of the \texttt{[SEP]} token\label{sec:ablation_sep}}
We incorporate \texttt{[SEP]} tokens to delimit different input segments, following standard practice in textual Transformers~\cite{Radford2018}.
Nonetheless, track $\tau$ is the only element in input $\mathcal{I} = \{\tau, \mathcal{P}\}$ with variable length and, thus, separating each proposal $p \in \mathcal{P}$ using \texttt{[SEP]} is not necessarily mandatory.
Still, Tab.~\ref{tab:ablation-sep} shows the advantages of retaining \texttt{[SEP]} for every proposal, as visual Transformers benefit from having additional registers to store, process, and retrieve global information\cite{Darcet2024}.

\section{Impact of Training \byteformer on In-Domain Data \label{sec:finetuning_busca}}
\byteformer aims to be as portable and generic as possible to facilitate its integration with any type of tracker by detection.
This is why we train it on the MOTSynth~\cite{Fabbri2021} synthetic dataset, without making any adjustments for specific trackers or scenarios.
Still, in-domain training from scratch is possible, as shown in Tab.~\ref{tab:ablation-finetune} (trained on the first half of MOT17-train and validated on the second half).
Despite being trained on less than 2 minutes of video, \byteformer still improves +0.3 MOTA over the baseline.
Nevertheless, BUSCA still benefits from additional data, such as the 100 sequences from MOTSynth used in the primary experiments.
Accordingly, to further boost the performance of \byteformer.

\begin{table}[t!]
\centering
\caption{\byteformer results on DanceTrack~\cite{Seidenschwarz2022}.
$^{\star}$ denotes reproduced results for GHOST~\cite{Seidenschwarz2022} using the publicly available official code.}
\resizebox{0.75\linewidth}{!}{%
\begin{NiceTabular}{@{}l|c@{\hskip 3pt}cc@{\hskip 3pt}cc@{\hskip 3pt}cc@{\hskip 3pt}cc@{\hskip 3pt}c@{}}
\CodeBefore
   \rowcolors{13}{buscasota}{}
\Body
\midrule \\[-13pt]
& \multicolumn{2}{c}{\textbf{HOTA} $\uparrow$} & \multicolumn{2}{c}{\textbf{IDF1} $\uparrow$} & \multicolumn{2}{c}{\textbf{MOTA} $\uparrow$} & \multicolumn{2}{c}{\textbf{DetA} $\uparrow$} & \multicolumn{2}{c}{\textbf{AssA} $\uparrow$} \\[-2pt]
\midrule
\midrule

CenterTrack~\cite{Zhou2020b} & 41.8 && 35.7  && 86.8 && 78.1 && 22.6\\
FairMOT~\cite{Zhang2021b} & 39.7 && 40.8 && 82.2 && 66.7 && 23.8 \\
QDTrack~\cite{Pang2021} & 54.2 && 50.4 && 87.7 && 80.1 && 36.8\\
TransTrack~\cite{Sun2020b} & 45.5 && 45.2 && 88.4 && 75.9 && 27.5 \\
TraDeS~\cite{Wu2021} & 43.3 && 41.2 && 86.2 && 74.5 && 25.4 \\
MOTR~\cite{Zeng2022} & 54.2 && 51.5 && 79.7 && 73.5 && 40.2 \\
GTR~\cite{Zhou2022b} & 48.0 && 50.3 && 84.7 && 72.5 && 31.9 \\
ByteTrack~\cite{Zhang2022} & 47.7 && 53.9 && 89.6 && 71.0 && 32.1 \\

GHOST~\cite{Seidenschwarz2022} & 56.7 && 57.7 && 91.3 && 81.1 && 39.8 \\[-1pt]

\midrule
\midrule

GHOST$^{\star}$~\cite{Seidenschwarz2022}  & 54.8 && 55.5 && 91.3 && 81.1 && 37.1 &  \\

\plusours{} \textbf{(ours)}  &  \rankA{55.5} &  \gooddelta{+0.7} & \rankA{56.1} & \gooddelta{+0.6} & \rankA{91.5} & \gooddelta{+0.2} & \rankA{81.4} & \gooddelta{+0.3} & \rankA{38.0} & \gooddelta{+0.9} \\[-2pt]
\midrule
\end{NiceTabular}}
\label{tab:dancetrack}
\end{table}

\section{Behaviour under Complex Motions\label{sec:dancetrack}}
\byteformer's simple-yet-effective design is also able to model complex motions, like the ones found in DanceTrack~\cite{sun2022dance}.
Among the five trackers we employed, only GHOST~\cite{Seidenschwarz2022} provides an official code for DanceTrack.
We thus take it as an example and show here its test results with and without \byteformer.
While the official code does not replicate the exact results in~\cite{Seidenschwarz2022}, \byteformer still yields improvements in DanceTrack, as shown in Tab.~\ref{tab:dancetrack}.

\begin{table}[t!]
\centering
\caption{\byteformer results on BDD100K~\cite{Yu2020c} validation set.
$^{\star}$ denotes reproduced results for GHOST~\cite{Seidenschwarz2022} using the publicly available official code.}
\resizebox{0.60\linewidth}{!}{%
\begin{NiceTabular}{@{}l|c@{\hskip 3pt}c@{\hskip 7pt}c@{\hskip 3pt}c@{\hskip 7pt}c@{\hskip 3pt}c@{}}
\CodeBefore
   \rowcolors{9}{}{buscasota}
\Body
\midrule \\[-13pt]
& \multicolumn{2}{c}{\textbf{mMOTA} $\uparrow$} & \multicolumn{2}{c}{\textbf{mHOTA} $\uparrow$} & \multicolumn{2}{c}{\textbf{mIDF1} $\uparrow$} \\[-2pt]
\midrule
\midrule

D2TT2D-100K~\cite{Yu2020c} & 25.9 && -- && 44.5 \\
MOTR~\cite{Zeng2022} & 32.0 && -- && 43.5 \\
QDTrack~\cite{Pang2021} & 36.3 && 41.7 && 51.5 \\
TETer~\cite{Liu2020} & 39.1 && -- && 53.3 \\
GHOST~\cite{Seidenschwarz2022} & 44.9 && 45.7 && 55.6 \\
ByteTrack~\cite{Zhang2022} & 45.2 && 45.4 && 54.6 \\[-1pt]

\midrule
\midrule

GHOST$^{\star}$~\cite{Seidenschwarz2022}  & 43.4 && 42.5 && 50.7  \\

\plusours{} \textbf{(ours)}  &  \rankA{43.7} &  \gooddelta{+0.3} & \rankA{43.1} & \gooddelta{+0.6} & \rankA{52.4} & \gooddelta{+1.7} \\[-2pt]
\midrule
\end{NiceTabular}}
\label{tab:bdd100k}
\end{table}

\section{Performance on Different Categories\label{sec:bdd100k}}
The performance improvement that \byteformer yields is not limited to people only.
In Tab.~\ref{tab:bdd100k}, we show how it can improve the mMOTA, mHOTA, and mIDF1 of GHOST~\cite{Seidenschwarz2022} (other tested baseline trackers do not provide an official implementation for the dataset) in BDD100K~\cite{Yu2020c}, which comprises eight different categories.
For this experiment, we freeze the appearance feature extractor and fine-tune \byteformer on SHIFT~\cite{Sun2022b}.

\section{\byteformer vs. Offline Processing} \label{sec:vsoffline}

\byteformer offers a comprehensive framework to persistently track objects missed by the detector in a \emph{fully online} manner (i.e., without modifying past tracking predictions or accessing future frames).
This feature renders it highly valuable for applications where the solution has to be immediately available with each incoming frame and cannot be changed at any later time.
Still, certain algorithms in the literature recover objects post hoc through offline post-processing techniques.
While these offline techniques are not directly comparable to \byteformer, their results serve as a possible insight into the theoretical upper bound for online methods like \byteformer.
Specifically, Tab.~\ref{tab:sota-byte-offline} showcases the offline mechanisms employed in StrongSORT++~\cite{du2023strongsort} (i.e., AFLink and GSI) and in ByteTrack~\cite{Zhang2022} (i.e., offline linear interpolation and per-sequence thresholds).

\begin{table*}[!t]
\setlength\extrarowheight{-3pt} 
\centering
\caption{State-of-the-art comparison on MOT17 and MOT20 test sets including offline versions of contemporary methods (colored gray). $^{\star}$ means that the offline interpolation and the per-sequence thresholds in ByteTrack~\cite{Zhang2022} are removed.
}
\resizebox{\linewidth}{!}{%
\begin{NiceTabular}{@{}l@{\hskip -10pt} c@{\hskip -8pt}l@{\hskip -2pt}c@{\hskip -7pt}l@{\hskip 0pt}c@{\hskip -4pt}l@{\hskip 2pt} | c@{\hskip -8pt}l@{\hskip -2pt} c@{\hskip -7pt}l@{\hskip 0pt}c@{\hskip -4pt}l@{\hskip 2pt}}
\CodeBefore
   \rowcolors{6-7,10}{buscasota}{}
   \rowcolors{9}{}{buscasota}
\Body
\hline \\[-6pt]
& \multicolumn{6}{c}{\textbf{MOT17}} & & \multicolumn{3}{c}{\textbf{MOT20}} \\

 & \textbf{MOTA} $\uparrow$ && \textbf{HOTA} $\uparrow$ && \textbf{IDF1} $\uparrow$ && \textbf{MOTA} $\uparrow$ && \textbf{HOTA} $\uparrow$ && \textbf{IDF1} $\uparrow$  & \\[-1pt]

\midrule
\midrule

StrongSORT~\cite{du2023strongsort} & 78.3 && 63.5 && 78.5 && 72.2 && 61.5 && 75.9 & \\
\RowStyle{\color{badgray}} \ + {\smaller{AFLink}} ({\smaller{StrongSORT+}}) & 78.3 & \baddelta{+0.0} & 63.7 & \baddelta{+0.2} & 79.0 & \baddelta{+0.5} & 72.2 & \baddelta{+0.0} & 61.6 & \baddelta{+0.1} & 76.3 & \baddelta{+0.4} \\ 
\RowStyle{\color{badgray}} \ + {\smaller{AFLink}}+{\smaller{GSI}} ({\smaller{StrongSORT++}}) & \rankB{79.6} & \baddelta{+1.3} & 64.4 & \baddelta{+0.9} & 79.5 & \baddelta{+1.0} & \rankB{73.8} & \baddelta{+1.6} & 62.6 & \baddelta{+1.1} & 77.0 & \baddelta{+1.1} \\ 
\plusours{} \textbf{(ours)} & \textbf{78.6} & \gooddelta{+0.3} & \textbf{63.9} & \gooddelta{+0.4} & \textbf{79.2} & \gooddelta{+0.7} & \textbf{72.7} & \gooddelta{+0.5} & \textbf{61.8} & \gooddelta{+0.3} & \textbf{76.3} & \gooddelta{+0.4} \\[-1pt]

\midrule
\midrule

ByteTrack$^{\star}$~\cite{Zhang2022} & \rankC{78.9} && \rankC{62.8} && \rankB{77.1} && \rankB{74.2} && 60.4 && \rankC{74.5} & \\ 
\RowStyle{\color{badgray}} \ + interp.+thresh. ({\smaller{ByteTrack}}) & \rankB{80.3} & \baddelta{+1.4} & 63.1 & \baddelta{+0.3} & 77.3 & \baddelta{+0.2} & \rankB{77.8} & \baddelta{+3.6} & 61.3 & \baddelta{+0.9} & 75.2 & \baddelta{+0.7}\\ 
\plusours{} \textbf{(ours)} & \textbf{79.3} & \gooddelta{+0.4} & \textbf{63.1} & \gooddelta{+0.3} & \textbf{77.7} & \gooddelta{+0.6} & \textbf{74.5} & \gooddelta{+0.3} & \textbf{60.5} & \gooddelta{+0.1} & \textbf{74.4} & \baddelta{\text{-}0.1} \\[-2pt]
\midrule

\end{NiceTabular}}
\label{tab:sota-byte-offline}
\end{table*}

AFLink, an appearance-free linking model that leverages spatiotemporal information to predict if two tracklets belong to the same object ID, provides a slight boost in HOTA and IDF1, albeit still inferior to \byteformer's.
With the addition of GSI, which employs Gaussian process regression~\cite{Williams1995} for bounding box interpolation, the achieved MOTA surpasses \byteformer by one point, highlighting the importance of handling extremely occluded objects.
This effect is further emphasized with ByteTrack's offline linear interpolation and tracking thresholds, which are adapted based on the evaluated test sequence.
Nevertheless, \byteformer's performance remains competitive, consistently enhancing the capabilities of TbD trackers in a fully online manner.

Lastly, despite being designed to enhance online multi-object trackers, \byteformer can also potentially improve batch-based and offline tracking algorithms.
Exhaustive analysis in this regard falls out of the scope of this paper.
However, for demonstration purposes, we show in Tab.~\ref{tab:benchmark-mot-offline} how integrating \byteformer with the interpolation-based offline version of~\cite{Zhang2022} results in +0.7/+0.2 MOTA/HOTA in the MOT17 validation set.

\begin{table}[t!]
\centering
\caption{Compatibility of \byteformer with offline algorithms.
$^{\star}$ means that the offline interpolation
and the per-sequence thresholds in ByteTrack~\cite{Zhang2022} are removed.}
\label{tab:benchmark-mot-offline}
\resizebox{0.70\linewidth}{!}{%
\begin{tabular}{@{}l|c@{\hskip 3pt}cc@{\hskip 3pt}cc@{\hskip 3pt}cc@{\hskip 3pt}c}
\toprule

& \multicolumn{2}{c}{\textbf{MOTA} $\uparrow$} & \multicolumn{2}{c}{\textbf{HOTA} $\uparrow$} & \multicolumn{2}{c}{\textbf{IDF1} $\uparrow$} & \multicolumn{2}{c}{\textbf{IDSW} $\downarrow$} \\
\hline
\hline

ByteTrack$^{\star}$  & 76.5 &  & 67.4 & & 79.4 &  & \rankA{165} &  \\
\plusours \textbf{(ours)}  & \rankA{77.1} &  \gooddelta{+0.6} & \rankA{67.6} & \gooddelta{+0.2} & \rankA{79.5} & \gooddelta{+0.1} & 166 & \baddelta{+1}  \\[-1pt]

\midrule
\midrule

ByteTrack  & 77.8 &  & 67.9 & & 79.9 &  & 168 &  \\
\plusours \textbf{(ours)}  & \rankA{78.5} &  \gooddelta{+0.7} & \rankA{68.1} & \gooddelta{+0.2} & \rankA{80.1} & \gooddelta{+0.2} & \rankA{166} & \gooddelta{\text{-}2}  \\[-2pt]

\bottomrule
\end{tabular}}
\end{table}

\section{Sequence-Wise Results\label{sec:detailed}}
To facilitate the comparison of \byteformer with other approaches on a finer-grained level, we show sequence-wise results for the test sets of MOT16~\cite{Milan2016a}, MOT17~\cite{Milan2016a}, and MOT20 \cite{Dendorfer2020} in Tabs.~\ref{tab:sota-mot16-detailed}, \ref{tab:sota-mot17-detailed}, and \ref{tab:sota-mot20-detailed}, respectively.

\newpage
\begin{table*}[!t]
\setlength\extrarowheight{-3pt}
\centering
\caption{Sequence-wise results on MOT16 test set. $^{\star}$ means that the offline interpolation and the per-sequence thresholds in ByteTrack~\cite{Zhang2022} are removed for fair comparison.
Private detections are used.
\vspace*{-2mm}
}
\resizebox{0.98\linewidth}{!}{%
\begin{NiceTabular}{@{}l@{\hskip -3pt} c@{\hskip 2pt}c@{\hskip 2pt}c@{\hskip 2pt} | c@{\hskip 2pt}c@{\hskip 2pt}c@{\hskip 2pt} | c@{\hskip 2pt}c@{\hskip 2pt}c@{\hskip 2pt} | c@{\hskip 2pt}c@{\hskip 2pt}c@{\hskip 2pt} | c@{\hskip 2pt}c@{\hskip 2pt}c@{\hskip 2pt} | c@{\hskip 2pt}c@{\hskip 2pt}c@{\hskip 2pt} | c@{\hskip 2pt}c@{\hskip 2pt}c@{\hskip 1pt}c@{}}
\hline \\[-6pt]
& \multicolumn{3}{c}{\textbf{MOT16-01}} &  \multicolumn{3}{c}{\textbf{MOT16-03}} & \multicolumn{3}{c}{\textbf{MOT16-06}} & \multicolumn{3}{c}{\textbf{MOT16-07}} & \multicolumn{3}{c}{\textbf{MOT16-08}} & \multicolumn{3}{c}{\textbf{MOT16-12}} & \multicolumn{3}{c}{\textbf{MOT16-14}} \\

 & {\smaller \smaller {\textbf{MOTA}}} & {\smaller \smaller {\textbf{HOTA}}} & {\smaller \smaller {\textbf{IDF1}}} & {\smaller \smaller {\textbf{MOTA}}} & {\smaller \smaller {\textbf{HOTA}}} & {\smaller \smaller {\textbf{IDF1}}} & {\smaller \smaller {\textbf{MOTA}}} & {\smaller \smaller {\textbf{HOTA}}} & {\smaller \smaller {\textbf{IDF1}}} & {\smaller \smaller {\textbf{MOTA}}} & {\smaller \smaller {\textbf{HOTA}}} & {\smaller \smaller {\textbf{IDF1}}} & {\smaller \smaller {\textbf{MOTA}}} & {\smaller \smaller {\textbf{HOTA}}} & {\smaller \smaller {\textbf{IDF1}}} & {\smaller \smaller {\textbf{MOTA}}} & {\smaller \smaller {\textbf{HOTA}}} & {\smaller \smaller {\textbf{IDF1}}} & {\smaller \smaller {\textbf{MOTA}}} & {\smaller \smaller {\textbf{HOTA}}} & {\smaller \smaller {\textbf{IDF1}}} &\\[-1pt]

\midrule
\midrule

\begin{tabular}{@{}c@{}}CenterTrack \\ \plusours{}\end{tabular} & 60.5 & 45.1 & 53.1 & 86.6 & 63.8 & 80.6 & 56.9 & 45.9 & 59.5 & 55.2 & 41.5 & 50.7 & 39.2 & 42.0 & 46.7 & 48.9 & 48.8 & 60.4 & 41.7 & 37.5 & 50.4 \\[-2pt]

\midrule

\begin{tabular}{@{}c@{}}TransCenter \\ \plusours{}\end{tabular} & 55.3 & 49.0 & 58.5 & 90.7 & 71.9 & 85.8 & 58.8 & 45.0 & 56.0 & 67.7 & 49.9 & 60.0 & 50.6 & 44.5 & 51.6 & 54.9 & 53.7 & 67.1 & 48.1 & 42.5 & 60.5 \\[-2pt]

\midrule

\begin{tabular}{@{}c@{}}GHOST \\ \plusours{}\end{tabular} & 60.2 & 52.1 & 61.6 & 91.5 & 71.2 & 87.9 & 61.6 & 51.7 & 62.7 & 71.4 & 51.4 & 60.1 & 56.4 & 49.8 & 57.2 & 57.0 & 56.5 & 68.0 & 57.2 & 48.4 & 65.9 \\[-2pt]

\midrule

\begin{tabular}{@{}c@{}}StrongSORT \\ \plusours{}\end{tabular} & 61.7 & 52.0 & 63.2 & 91.4 & 72.0 & 89.0 & 62.3 & 51.6 & 63.5 & 71.3 & 55.2 & 69.7 & 56.5 & 48.8 & 56.3 & 62.9 & 59.8 & 73.1 & 54.3 & 48.6 & 68.1 \\[-2pt]

\midrule

\begin{tabular}{@{}c@{}}ByteTrack$^{\star}$ \\ \plusours{}\end{tabular} & 61.6 & 49.2 & 57.8 & 91.6 & 72.5 & 90.3 & 61.5 & 47.1 & 57.3 & 72.1 & 49.4 & 59.5 & 55.3 & 46.4 & 51.8 & 64.3 & 58.6 & 71.5 & 53.8 & 45.5 & 63.6 \\[-2pt]
\midrule

\end{NiceTabular}}
\label{tab:sota-mot16-detailed}
\end{table*}

\begin{table*}[!t]
\setlength\extrarowheight{-3pt}
\centering
\caption{Sequence-wise results on MOT17 test set. $^{\star}$ means that the offline interpolation and the per-sequence thresholds in ByteTrack~\cite{Zhang2022} are removed for fair comparison.
Private detections are used.
\vspace*{-2mm}
}
\resizebox{0.98\linewidth}{!}{%
\begin{NiceTabular}{@{}l@{\hskip -3pt} c@{\hskip 2pt}c@{\hskip 2pt}c@{\hskip 2pt} | c@{\hskip 2pt}c@{\hskip 2pt}c@{\hskip 2pt} | c@{\hskip 2pt}c@{\hskip 2pt}c@{\hskip 2pt} | c@{\hskip 2pt}c@{\hskip 2pt}c@{\hskip 2pt} | c@{\hskip 2pt}c@{\hskip 2pt}c@{\hskip 2pt} | c@{\hskip 2pt}c@{\hskip 2pt}c@{\hskip 2pt} | c@{\hskip 2pt}c@{\hskip 2pt}c@{\hskip 1pt}c@{}} 
\hline \\[-6pt]
& \multicolumn{3}{c}{\textbf{MOT17-01}} &  \multicolumn{3}{c}{\textbf{MOT17-03}} & \multicolumn{3}{c}{\textbf{MOT17-06}} & \multicolumn{3}{c}{\textbf{MOT17-07}} & \multicolumn{3}{c}{\textbf{MOT17-08}} & \multicolumn{3}{c}{\textbf{MOT17-12}} & \multicolumn{3}{c}{\textbf{MOT17-14}} \\

 & {\smaller \smaller {\textbf{MOTA}}} & {\smaller \smaller {\textbf{HOTA}}} & {\smaller \smaller {\textbf{IDF1}}} & {\smaller \smaller {\textbf{MOTA}}} & {\smaller \smaller {\textbf{HOTA}}} & {\smaller \smaller {\textbf{IDF1}}} & {\smaller \smaller {\textbf{MOTA}}} & {\smaller \smaller {\textbf{HOTA}}} & {\smaller \smaller {\textbf{IDF1}}} & {\smaller \smaller {\textbf{MOTA}}} & {\smaller \smaller {\textbf{HOTA}}} & {\smaller \smaller {\textbf{IDF1}}} & {\smaller \smaller {\textbf{MOTA}}} & {\smaller \smaller {\textbf{HOTA}}} & {\smaller \smaller {\textbf{IDF1}}} & {\smaller \smaller {\textbf{MOTA}}} & {\smaller \smaller {\textbf{HOTA}}} & {\smaller \smaller {\textbf{IDF1}}} & {\smaller \smaller {\textbf{MOTA}}} & {\smaller \smaller {\textbf{HOTA}}} & {\smaller \smaller {\textbf{IDF1}}} &\\[-1pt]

\midrule
\midrule

\begin{tabular}{@{}c@{}}CenterTrack \\ \plusours{}\end{tabular} & 60.3 & 45.1 & 53.0 & 86.8 & 64.0 & 80.7 & 57.9 & 45.9 & 59.3 & 55.4 & 41.0 & 50.1 & 32.0 & 37.6 & 40.4 & 48.0 & 48.2 & 59.4 & 41.9 & 37.5 & 50.4 \\[-2pt]

\midrule

\begin{tabular}{@{}c@{}}TransCenter \\ \plusours{}\end{tabular} & 57.3 & 49.5 & 59.0 & 90.9 & 72.2 & 85.9 & 60.7 & 45.1 & 55.8 & 68.1 & 49.4 & 59.3 & 58.0 & 43.1 & 50.6 & 54.4 & 53.1 & 66.2 & 48.1 & 42.5 & 60.5 \\[-2pt]

\midrule

\begin{tabular}{@{}c@{}}GHOST \\ \plusours{}\end{tabular} & 60.4 & 52.3 & 62.3 & 92.5 & 71.3 & 88.2 & 63.3 & 51.9 & 62.4 & 71.5 & 51.7 & 60.4 & 61.1 & 47.4 & 54.2 & 56.8 & 55.8 & 66.7 & 57.2 & 48.4 & 65.9 \\[-2pt]

\midrule

\begin{tabular}{@{}c@{}}StrongSORT \\ \plusours{}\end{tabular} & 61.8 & 52.1 & 63.2 & 92.3 & 72.3 & 89.5 & 62.9 & 51.5 & 63.0 & 71.0 & 55.0 & 69.3 & 58.9 & 46.4 & 54.2 & 62.0 & 59.1 & 72.4 & 54.3 & 48.6 & 68.1 \\[-2pt]

\midrule

\begin{tabular}{@{}c@{}}ByteTrack$^{\star}$ \\ \plusours{}\end{tabular} & 61.5 & 49.3 & 57.8 & 92.6 & 72.8 & 90.8 & 62.3 & 47.0 & 56.9 & 73.1 & 49.7 & 59.9 & 61.8 & 44.7 & 50.4 & 63.3 & 57.8 & 70.6 & 53.8 & 45.5 & 63.6 \\[-2pt]
\midrule

\end{NiceTabular}}
\label{tab:sota-mot17-detailed}
\end{table*}

\begin{table*}[!t]
\setlength\extrarowheight{-3pt} 
\centering
\caption{Sequence-wise results on MOT20 test set. $^{\star}$ means that the offline interpolation and the per-sequence thresholds in ByteTrack~\cite{Zhang2022} are removed for fair comparison.
Private detections are used.
\vspace*{-2mm}
}
\resizebox{0.65\linewidth}{!}{%
\begin{NiceTabular}{@{}l@{\hskip -3pt} c@{\hskip 2pt}c@{\hskip 6pt}c@{\hskip 2pt} | c@{\hskip 2pt}c@{\hskip 6pt}c@{\hskip 2pt} | c@{\hskip 2pt}c@{\hskip 6pt}c@{\hskip 2pt} | c@{\hskip 2pt}c@{\hskip 6pt}c@{\hskip 1pt}c@{}}
\hline \\[-6pt] 
& \multicolumn{3}{c}{\textbf{MOT20-04}} &  \multicolumn{3}{c}{\textbf{MOT20-06}} & \multicolumn{3}{c}{\textbf{MOT20-07}} & \multicolumn{3}{c}{\textbf{MOT20-08}} \\

 & {\smaller \smaller {\textbf{MOTA}}} & {\smaller \smaller {\textbf{HOTA}}} & {\smaller \smaller {\textbf{IDF1}}} & {\smaller \smaller {\textbf{MOTA}}} & {\smaller \smaller {\textbf{HOTA}}} & {\smaller \smaller {\textbf{IDF1}}} & {\smaller \smaller {\textbf{MOTA}}} & {\smaller \smaller {\textbf{HOTA}}} & {\smaller \smaller {\textbf{IDF1}}} & {\smaller \smaller {\textbf{MOTA}}} & {\smaller \smaller {\textbf{HOTA}}} & {\smaller \smaller {\textbf{IDF1}}} &\\[-1pt]

\midrule
\midrule

\begin{tabular}{@{}c@{}}CenterTrack \\ \plusours{}\end{tabular} & 57.1 & 48.3 & 65.6 & 42.8 & 37.7 & 48.1 & 73.9 & 56.1 & 69.3 & 23.8 & 34.5 & 44.2 \\[-2pt]

\midrule

\begin{tabular}{@{}c@{}}TransCenter \\ \plusours{}\end{tabular} & 84.9 & 66.1 & 82.2 & 61.1 & 49.0 & 58.6 & 77.7 & 59.3 & 71.1 & 55.6 & 44.9 & 57.2 \\[-2pt]

\midrule

\begin{tabular}{@{}c@{}}GHOST \\ \plusours{}\end{tabular} & 87.3 & 69.2 & 84.3 & 59.8 & 50.1 & 62.2 & 81.8 & 64.5 & 75.6 & 49.6 & 44.3 & 56.9 \\[-2pt]

\midrule

\begin{tabular}{@{}c@{}}StrongSORT \\ \plusours{}\end{tabular} & 87.1 & 69.4 & 84.7 & 56.8 & 51.1 & 65.2 & 81.9 & 67.6 & 79.8 & 45.3 & 43.5 & 56.1 \\[-2pt]

\midrule

\begin{tabular}{@{}c@{}}ByteTrack$^{\star}$ \\ \plusours{}\end{tabular} & 86.8 & 67.8 & 83.6 & 61.4 & 50.7 & 62.4 & 81.3 & 62.9 & 73.2 & 50.3 & 43.7 & 55.7 \\[-2pt]
\midrule

\end{NiceTabular}}
\label{tab:sota-mot20-detailed}
\end{table*}

\begin{figure*}[ht]
\centering
\includegraphics[width=0.95\linewidth]{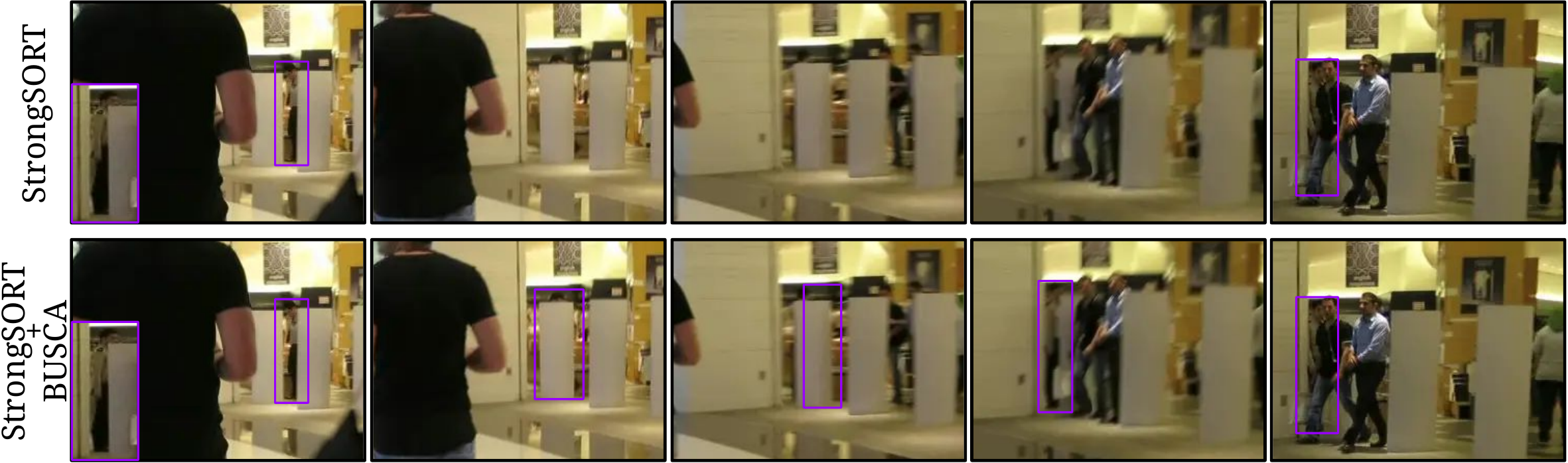}
\vspace*{3mm}
\includegraphics[width=0.95\linewidth]{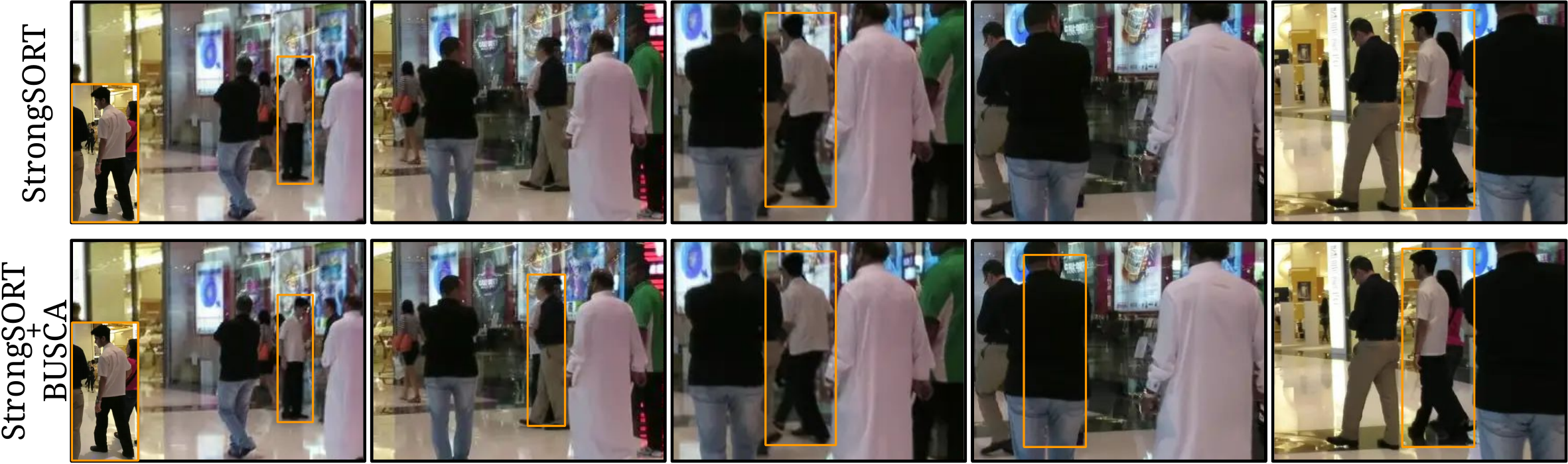}
\caption{Qualitative results showing the benefit of integrating \byteformer into StrongSORT~\cite{du2023strongsort}. Colors represent object identities. Results are shown for only one subject to ease the visualization.
}
\label{fig:quali_strongsort}
\end{figure*}
\vspace{-5cm}
\begin{figure*}[ht]
\centering
\includegraphics[width=0.95\linewidth]{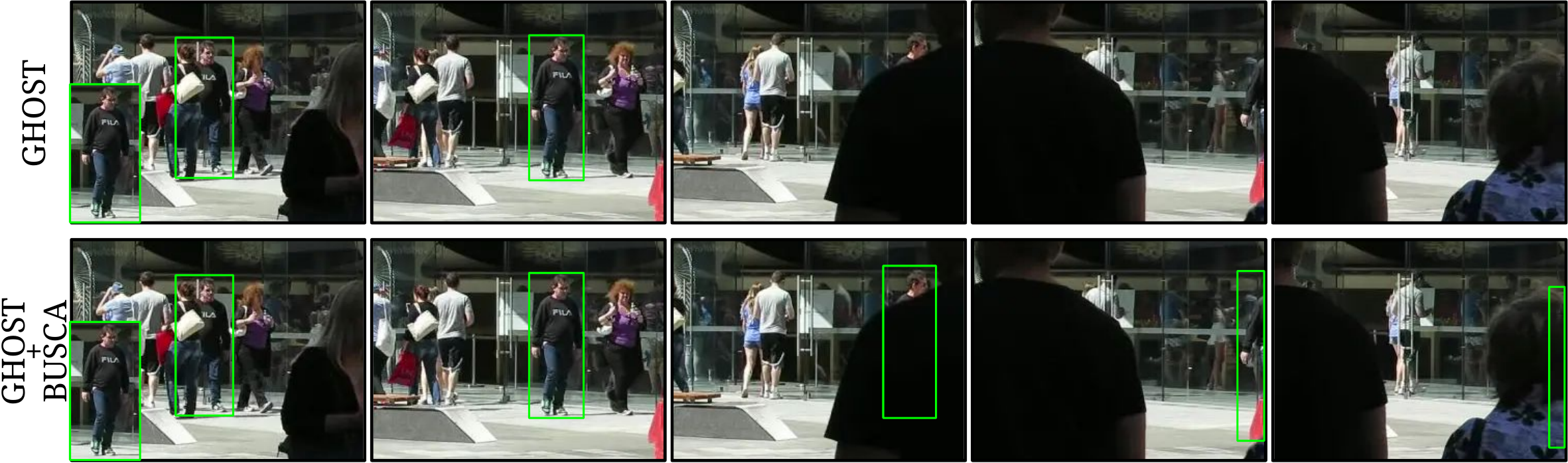}
\vspace*{3mm}\\
\includegraphics[width=0.95\linewidth]{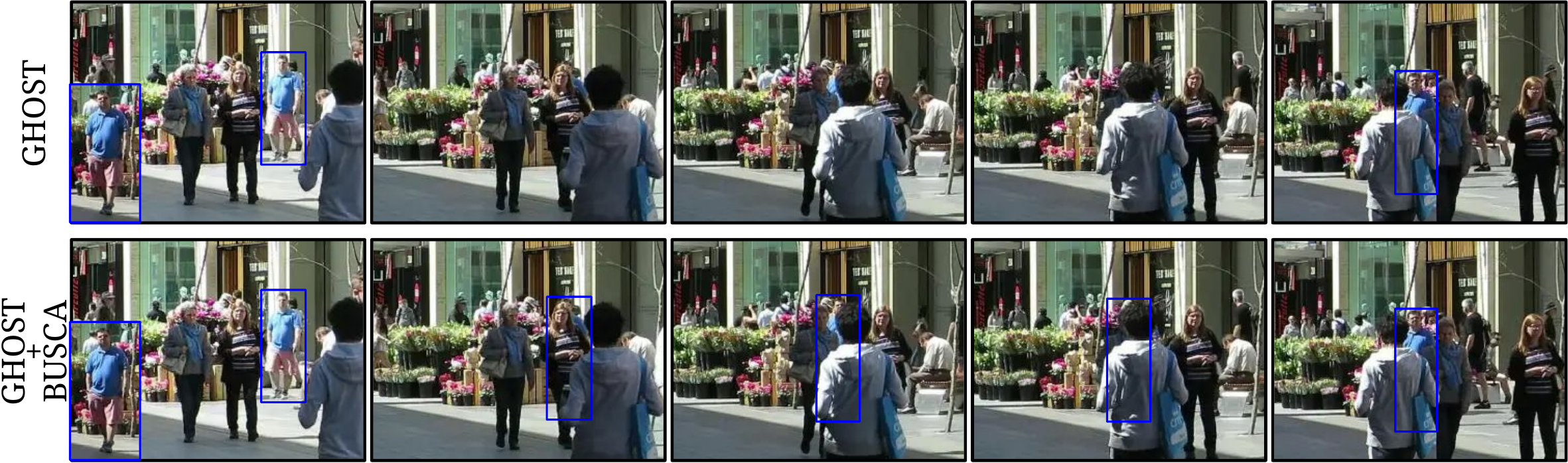}
\caption{Qualitative results showing the benefit of integrating \byteformer into GHOST~\cite{Seidenschwarz2022}. Colors represent object identities. Results are shown for only one subject to ease the visualization.
}
\label{fig:quali_ghost}
\end{figure*}

\begin{figure*}[ht]
\centering
\includegraphics[width=0.95\linewidth]{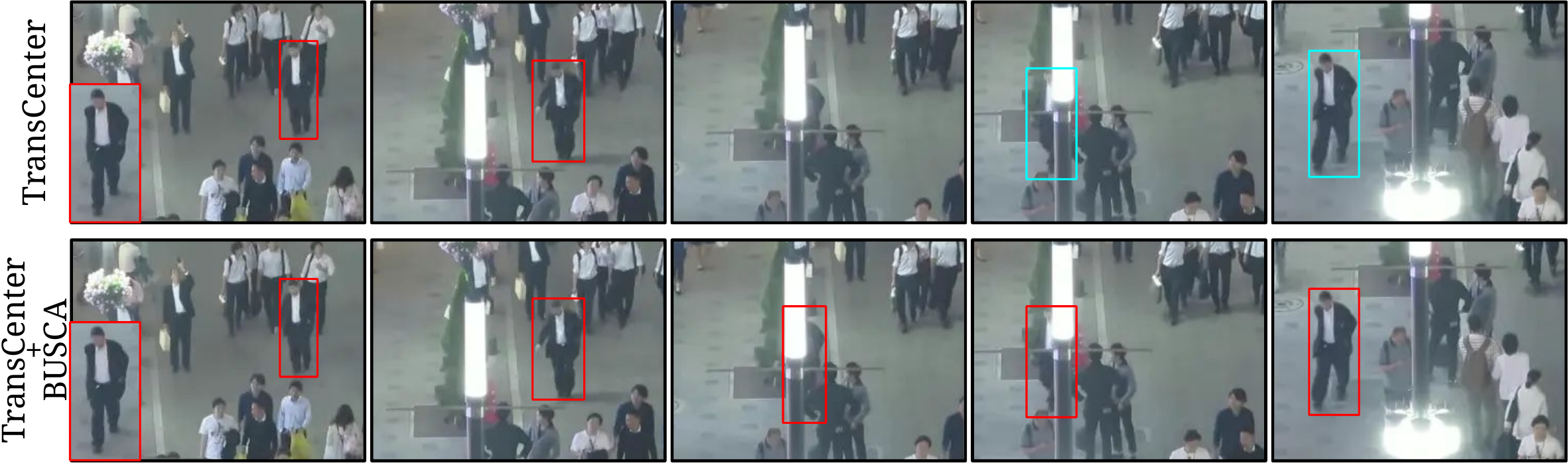}
\vspace*{3mm}\\
\includegraphics[width=0.95\linewidth]{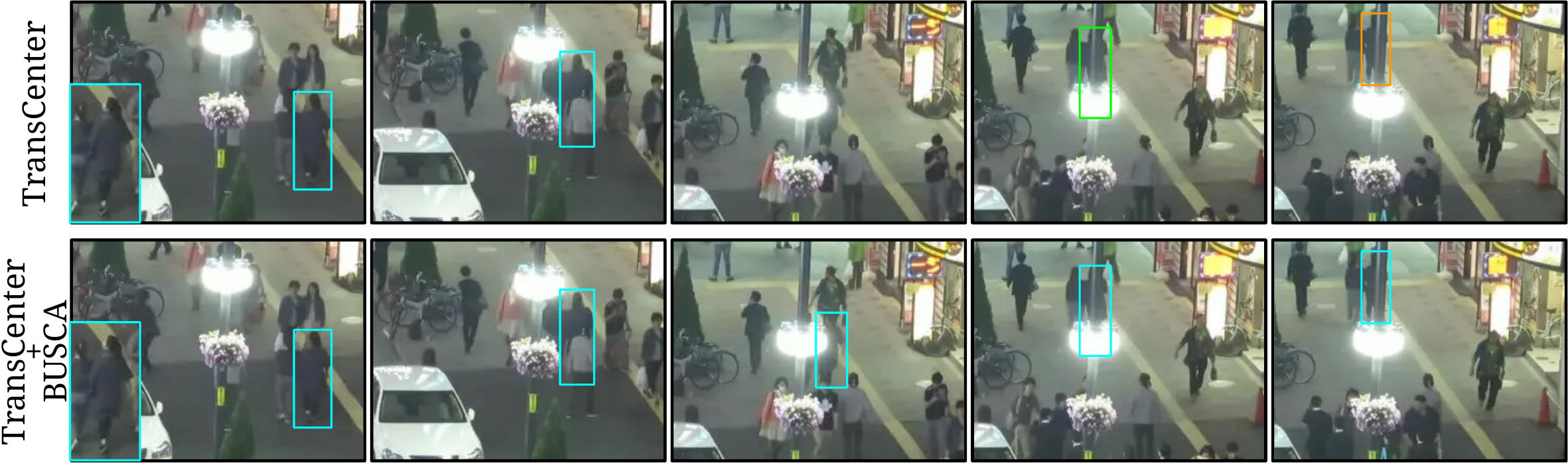}
\caption{Qualitative results showing the benefit of integrating \byteformer into TransCenter~\cite{Xu2023}. Colors represent object identities. Results are shown for only one subject to ease the visualization.
}
\label{fig:quali_transcenter}
\end{figure*}

\begin{figure*}[ht]
\centering
\includegraphics[width=0.95\linewidth]{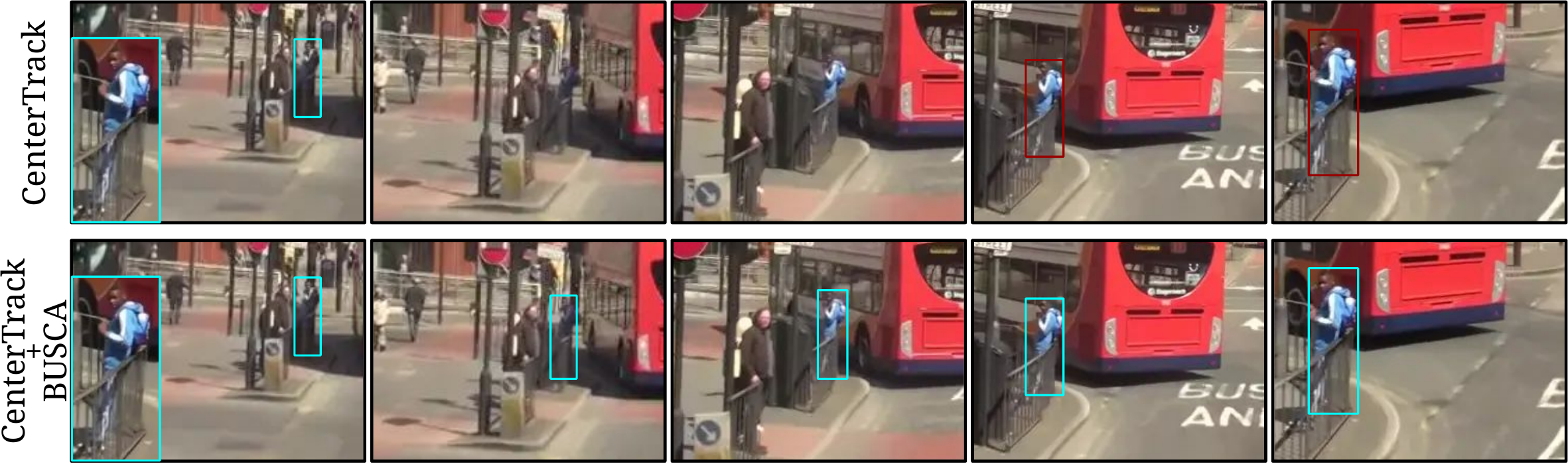}
\vspace*{3mm}\\
\includegraphics[width=0.95\linewidth]{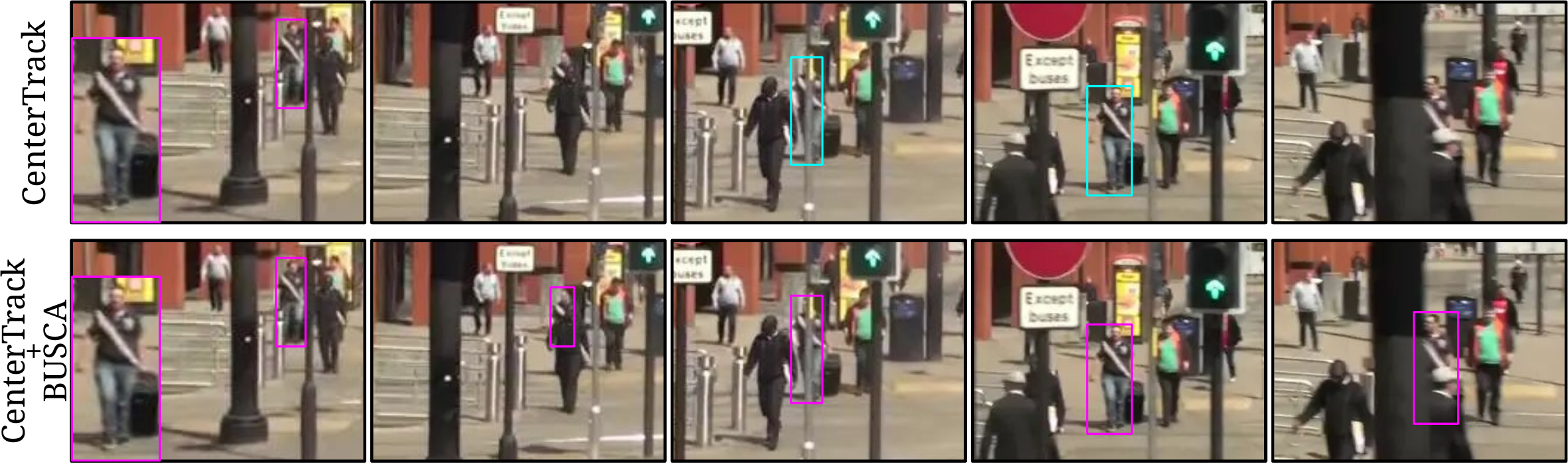}
\caption{Qualitative results showing the benefit of integrating \byteformer into CenterTrack~\cite{Zhou2020b}. Colors represent object identities. Results are shown for only one subject to ease the visualization.
}
\label{fig:quali_centertrack}
\end{figure*}

\clearpage

\end{document}